\pdfoutput=1

\documentclass[11pt]{article}

\usepackage[]{acl}

\usepackage{times}
\usepackage{latexsym}
\usepackage{graphicx}
\usepackage{arydshln}
\usepackage{amsmath}
\usepackage{subcaption}
\usepackage{multirow}
\usepackage{hyperref}
\usepackage{booktabs}
\usepackage{float}
\usepackage[T1]{fontenc}
\newcommand{\st}{Statement-Tuning}
\newcommand{\mst}{Multilingual Statement-Tuning}

\usepackage[utf8]{inputenc}

\usepackage{microtype}

\title{Statement-Tuning Enables Efficient Cross-lingual Generalization in Encoder-only Models}

\author{Ahmed Elshabrawy\textsuperscript{\textnormal{1}}\hspace{0.4cm}
Thanh-Nhi Nguyen \textsuperscript{\thanks{* Equal contribution for work done during internship done at MBZUAI}*\textnormal{2, 3}} \hspace{0.4cm} Yeeun Kang \textsuperscript{*\textnormal{4}} \hspace{0.4cm} Lihan Feng \textsuperscript{*\textnormal{5}}   \\
\bf Annant Jain \textsuperscript{*\textnormal{6}} \hspace{0.4cm} Faadil A. Shaikh \textsuperscript{*} \hspace{0.4cm} Jonibek Mansurov \textsuperscript{\textnormal{1}} \hspace{0.4cm} Mohamed Fazli Imam \textsuperscript{\textnormal{1}} \\
\bf Jesus-German Ortiz-Barajas \textsuperscript{\textnormal{1}} \hspace{0.4cm} Rendi Chevi \textsuperscript{\textnormal{1}} \hspace{0.4cm} Alham Fikri Aji\textsuperscript{\textnormal{1}} \\
\textsuperscript{1} MBZUAI \hspace{0.4cm} \textsuperscript{2} UIT, Vietnam \hspace{0.4cm} \textsuperscript{3} VNU-HCM  \\
\textsuperscript{4} Yale University \hspace{0.4cm} \textsuperscript{5} NYU Shanghai \hspace{0.4cm}\textsuperscript{6} IIT Bombay \\
\small{\texttt{ahmed.elshabrawy@mbzuai.ac.ae}}}

\begin{document}
\maketitle
\begin{abstract}
Large Language Models (LLMs) excel in zero-shot and few-shot tasks, but achieving similar performance with encoder-only models like BERT and RoBERTa has been challenging due to their architecture. However, encoders offer advantages such as lower computational and memory costs. Recent work adapts them for zero-shot generalization using Statement Tuning, which reformulates tasks into finite templates. We extend this approach to multilingual NLP, exploring whether encoders can achieve zero-shot cross-lingual generalization and serve as efficient alternatives to memory-intensive LLMs for low-resource languages. Our results show that state-of-the-art encoder models generalize well across languages, rivaling multilingual LLMs while being more efficient. We also analyze multilingual Statement Tuning dataset design, efficiency gains, and language-specific generalization, contributing to more inclusive and resource-efficient NLP models.
We release our code and models\footnote{\href{https://github.com/mbzuai-nlp/Multilingual-ST}{https://github.com/mbzuai-nlp/Multilingual-ST}}.
\end{abstract}

\section{Introduction}
Large Language Models (LLMs) have shown great capabilities in zero-shot and few-shot settings \cite{radford-etal-2019-multitask, brown-etal-2020-fewshot, artetxe-etal-2022-efficient}. However, these capabilities are often more difficult to observe in encoder-only models like BERT \cite{devlin-etal-2019-bert} and RoBERTa \citep{liu2019roberta} due to their architectural design. These models are typically pretrained with a Masked Language Modeling (MLM) objective and are fine-tuned by adding task-specific layers to enable their usage on a downstream task. These task-specific layers block the extension of these models to new tasks in a few-shot or zero-shot manner. 

Despite these difficulties, applying encoder models for zero-shot task generalization offers several advantages. First, encoder models are typically more lightweight than LLMs and therefore require less computational power and memory. Further, encoder models excel at generating contextual embeddings that better capture semantic meaning. For instance, recent work \citep{qorib-etal-2024-decoder} demonstrated that decoder-only LLMs perform worse on word meaning comprehension than encoder-only models. Encoder-only models are also more efficient in inference for tasks such as sequence labeling due to their architecture \citep{soltan-etal-2023-recipes}. 

\begin{figure}
    \centering
    \includegraphics[width=\linewidth]{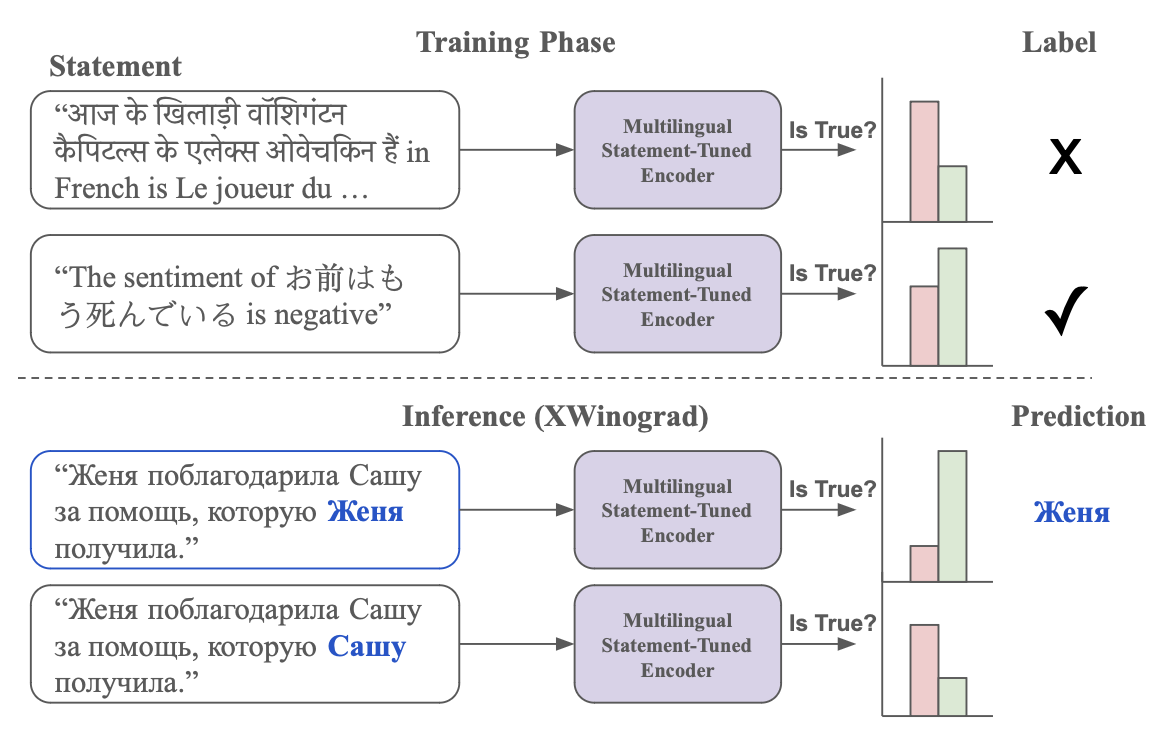}
    \caption{Encoder models trained with~\mst~ can generalize across new task and unseen languages during finetuning.}
    \label{fig:mst-demo}
\end{figure}

To enable encoder model usage in zero-shot task generalization, \citet{elshabrawy2024enablingnaturalzeroshotprompting} introduced Statement-Tuning. This technique converts tasks into a set of statements with finite templates, training on an encoder-only model, RoBERTa, to discriminate between potential statements and derive final results. This method demonstrates the feasibility of using encoder models, typically specialized for specific tasks, to handle various unseen Natural Language Understanding (NLU) tasks, similar to zero-shot prompting in decoder models. It shows competitive performance compared to large language models (LLMs) with significantly fewer parameters and training data, highlighting its potential for zero-shot learning. However, the original approach focused only on English, raising questions about its applicability in multilingual settings and its ability to generalize to new tasks and languages.

In this work, we aim to explore the adaptation of \st~ for multilingual NLP tasks. Specifically, we investigate whether encoder models can achieve zero-shot cross-lingual generalization similar to decoder-only LLMs \cite{muennighoff-etal-2023-crosslingual}. Given the multilingual setting, it is crucial to emphasize the importance of low-compute solutions. Speakers and users of low-resource languages often lack the computational resources necessary to utilize memory-intensive LLMs. Therefore, our method, which leverages efficient encoder models, is particularly important as it offers a more accessible and inclusive approach to zero-shot text classification in these contexts \cite{ruder2022statemultilingualai}.

Hence our contributions are as follows:

\begin{itemize}
    \itemsep0em
    \item We enable zero-shot generalization to unseen tasks and languages for encoder-only models, a capability typically limited to decoder-based models.  
    \item Our benchmarking shows that state-of-the-art multilingual encoder-only models match LLMs in performance while being more efficient.  
    \item We analyze multilingual \st~ dataset design, including language diversity and translated prompt templates.  
    \item We investigate when and how multilingual \st~ generalizes effectively across languages.  
    \item We compare \st~ inference speed and memory efficiency against generative models, showing significant advantages.  


\end{itemize}



\section{Related Work}
\paragraph{Zero-shot and Few-shot Approaches Using Encoder-Only Models}
There have been several works exploring prompt-based approaches to enable few-shot and zero-shot generalization in Encoder-only Models. Finetuning on cloze templates or label discrimination \cite{schick-schutze-2021-exploiting, gao-etal-2021-making} effectively utilizes encoder models for few-shot learning. Cloze templates have also been shown to fare better than regular finetuning in cross-lingual few-shot transfer to unseen languages from higher resourced languages \cite{zhao-schutze-2021-discrete, tu-etal-2022-prompt, ma-etal-2023-prompt, ma-etal-2024-topro}.

However, to enable zero-shot task learning a reformulation of text classification tasks is necessary. \citet{yin-etal-2019-benchmarking} introduce the reformulation of any zero-shot text classification in the form of entailment where statements can be formed from a series of choices and the correct choice is seen to be an entailment. \citet{xu-etal-2023-universal} use DeBERTa to show that the entailment formulation of zero-shot classification can be observed to be more effective than the generative approach employed by LLMs. 

Finally, \citet{elshabrawy2024enablingnaturalzeroshotprompting} propose \st~to show that through template-based data augmentation, much smaller RoBERTa models can be finetuned on limited data to match or even exceed the zero-shot NLU capabilities of several LLMs of up to 70B parameters on monolingual classification tasks. While \citet{elshabrawy2024enablingnaturalzeroshotprompting} focused only on English, we studied whether \st~method is possible in other languages. Additionally, we explore the efficiency of the approach in more detail and offer insight on the effect of pretraining data on the performance of \st. 



\paragraph{Zero-shot Prompting and Multitask Tuning}
While LLMs were shown to perform well on few-shot generalization \cite{brown-etal-2020-fewshot}, they showed less successful performance on zero-shot generalization. To tackle this issue, instruction tuning was proposed. Instruction tuning refers to fine-tuning language models on a collection of datasets described via instructions \cite{DBLP:conf/iclr/WeiBZGYLDDL22}. Their model, Finetuned Language Net (FLAN), a decoder-only model of 137B parameters fine-tuned on more than 60 NLP datasets expressed via natural language instructions, proved effective in improving the zero-shot performance of models.

They also showed that increasing the number of tasks involved in instruction tuning improves unseen task generalization performance and asserted that the benefits of instruction tuning are emerging abilities of language models (i.e., they emerge with sufficient scale). Subsequent work by \citet{sanh2022multitask} explored instruction tuning with T5 encoder-decoder models and proposed the T0 models and datasets. They fine-tuned T5 models of 3B and 11B parameters, which were smaller than the FLAN model but still within the billions-of-parameters range. Their findings established that with a more diverse prompt setup and an encoder-decoder model like T5, language models could achieve good performance with instruction tuning.

\citet{chung2022scaling} found that instruction tuning is effective across a variety of model classes, such as PaLM, T5, and U-PaLM, as well as different prompting setups including zero-shot, few-shot, and chain-of-thought. Their models, FLAN-T5, ranged from 80M to 11B parameters and showed better performance than prior T5 checkpoints. Meanwhile, \citet{mishra-etal-2022-cross, wang-etal-2022-super, honovich-etal-2023-unnatural, wang-etal-2023-self-instruct} also proposed large-scale natural language instruction datasets.

These methods fine-tune large models on constructed datasets with various task prompts, achieving strong zero-shot results. However, effective instruction-tuned models often require billions of parameters \cite{zhang2024instruction}, limiting their application to smaller models. \citet{ye-etal-2022-zerogen} aim to distill this zero-shot ability in a smaller model like an LSTM through synthetic data creation using an LLM, but they create task-specific models rather than a single smaller model that is capable of generalizing.

Our work demonstrates achieving similar or superior generalization of LLMs using a single smaller MLM with less training data. Furthermore, our work expands on previous efforts by exploring encoder models, which contributes to parallel understanding when combined with works on decoder models \cite{DBLP:conf/iclr/WeiBZGYLDDL22} and encoder-decoder models \cite{sanh2022multitask}.

\begin{table*}[ht!]
\centering
\resizebox{0.8\linewidth}{!}{
\begin{tabular}{lcccccc}
\toprule
\textbf{Model} & \textbf{Parameters} & \textbf{XCOPA} & \textbf{XNLI} & \textbf{XStoryCloze} & \textbf{XWinoGrad} \\
\midrule
Qwen2 & 72B & 67.84 & \textcolor{lightgray}{\textbf{42.10}} & \textcolor{lightgray}{\textbf{66.70}} & 84.02 \\
Llama3.1 & 70B & \textcolor{lightgray}{\textbf{62.24}} & \textcolor{lightgray}{\textbf{41.68}} & \textcolor{lightgray}{\textbf{68.32}} & 82.69 \\
Gemma 2 & 9B & 66.29 & \textcolor{lightgray}{\textbf{46.50}} & \textcolor{lightgray}{\textbf{67.41}} & 83.93 \\
Llama3.1 & 8B & \textcolor{lightgray}{\textbf{60.29}} & \textcolor{lightgray}{\textbf{44.39}} & \textcolor{lightgray}{\textbf{61.60}} & 80.49 \\
Aya 23 & 8B & \textcolor{lightgray}{\textbf{54.60}} & \textcolor{lightgray}{\textbf{42.44}} & \textcolor{lightgray}{\textbf{60.36}} & 69.36 \\
Aya 23 & 35B & \textcolor{lightgray}{\textbf{57.24}} & \textcolor{lightgray}{\textbf{44.09}} & \textcolor{lightgray}{\textbf{63.65}} & 72.69 \\
Gemma 2 & 27B & 68.65 & \textcolor{lightgray}{\textbf{45.41}} & \textcolor{lightgray}{\textbf{69.76}} & 85.26 \\
Gemma 2 & 2B & \textcolor{lightgray}{\textbf{53.15}} & \textcolor{lightgray}{\textbf{34.08}} & \textcolor{lightgray}{\textbf{50.76}} & 59.27 \\
Qwen2 & 1.5B & \textcolor{lightgray}{\textbf{53.44}} & \textcolor{lightgray}{\textbf{34.73}} & \textcolor{lightgray}{\textbf{51.87}} & 66.94 \\
Qwen2 & 500M & \textcolor{lightgray}{\textbf{53.13}} & \textcolor{lightgray}{\textbf{33.58}} & \textcolor{lightgray}{\textbf{50.05}} & 58.08 \\
\midrule
mBERT(base) & 110M & 52.47 & 34.51 & 48.30 & 50.68 \\
XLMR-base & 250M & 56.69 & 35.33 & 60.71 & 51.34 \\
XLMR-large & 560M & 64.36 & 45.76 & 78.78 & 54.26 \\
\textbf{mDeBERTa (Best)} & 276M & 65.52$_{(1.64)}$ & 47.84$_{(1.65)}$ & 73.53$_{(1.25)}$ & 54.75$_{(1.24)}$ \\
\bottomrule
\end{tabular}
}
\caption{\textbf{Accuracy of the multilingual decoder and encoder models finetuned on the same data mixture.} on XCOPA, XNLI, XStoryCloze, and XWinoGrad tasks. Results in grey highlight performances that are below the best-performing encoder model, mDeBERTa (276M). Additionally, we report the average standard deviation across languages over 3 training runs only for mDeBERTa to quantify the random deviation due to \st~training.}
\label{fig:decoder-baselines}
\end{table*}

\section{Method: Multilingual \st
}
In this section, we outline the steps involved in \st. 

\subsection{Multilingual Task Verbalization}

First, using templates as shown in Figure~\ref{fig:template-example}, tasks are verbalized in natural language statements. We then train the statement discriminator to classify these statements as true or false. 

\begin{figure}[h]
    \centering
    \fbox{
        \begin{minipage}{0.9\linewidth}
            \centering
            \texttt{"\{\{target\_word\}\}" means the same in "\{\{context\_1\}\}" and "\{\{context\_2\}\}"}
        \end{minipage}
    }
    \caption{Example of a statement template used during task verbalization for sentiment analysis.}
    \label{fig:template-example}
\end{figure}

As \citet{elshabrawy2024enablingnaturalzeroshotprompting} propose, any discriminative task with a finite set of targets can be verbalized into a finite set of natural language statements, one for each label. Similar to prompting, each task has its own statement templates (outlined in Appendix~\ref{sec:appendixa}). The truth label for training purposes on each statement depends on whether the statement contains the correct target label or not. 


\subsection{Statement Fine-Tuning Setup}
\label{sft-setup}
To create the training dataset for statement fine-tuning, we exhaustively generate statements across 9 NLU tasks using many varied statement templates per dataset. For a detailed breakdown of the datasets used and what tasks they cover, refer to Appendix~\ref{sec:training-tasks}.

The rule for task selection generally follows the structure in \citet{elshabrawy2024enablingnaturalzeroshotprompting}, except for adding the machine-translation task. For each task, we randomly choose 1500 rows of training data for each language, with a balance of labels (controlling for positive and negative examples i.e. 750 examples for each label). This ensures the encoder models have sufficient data to train and explore their potential to be the generalizers. For selected low-resource languages, the total amount of data may be less than 1500; in that case, we choose all of the specific data for training. Data from the machine-translation task is added to enhance the generality of low-resource language in the tasks lacking data and models' cross-lingual ability. The rest of the tasks are selected either because they are often addressed by using LLMs or because we hypothesize they may enhance models' language understanding.

Our compilation of multilingual datasets amounts to 25 languages, both high- and low-resource. We include the full list of languages and additional language-specific information in the Appendix \ref{sec:appendixi}.

We explore the different number of languages including in the dataset and the language of the statement templates. We fine-tune different multilingual encoder-only models, mBERT \cite{DBLP:journals/corr/abs-1810-04805}, mDeBERTa \cite{he2021deberta}, XLM-R base and large \cite{conneau-etal-2020-unsupervised} with a binary sequence classification head to predict the truth value of the statements. By fine-tuning the model across diverse tasks, languages, and templates, we hypothesize that the model should be able to generalize across unseen templates, unseen tasks, as well as unseen languages, as long as it can be transferred into a True/False statement, and the "unseen" languages are at least seen during the pre-training stage. Additionally, for mDeBERTa trained with an 11-language \st~ dataset, we report the average and standard deviation over 3 different training runs, to account for randomness and report it as such where appropriate. We were unable to perform this over all ablations due to the scale of the experimentation and limited computational and time resources. 

\subsection{Zero-Shot Inference}

To perform inference on statement-tuned models, we convert testing tasks into declarative statements. Specifically, we generate a statement for each possible label and predict its probability of being true. The final label for a given task is the statement with the highest probability. To ensure robustness across different phrasings, we experiment with various templates for each task during both training and evaluation.



\begin{figure}
    \centering
    \includegraphics[width=\linewidth]{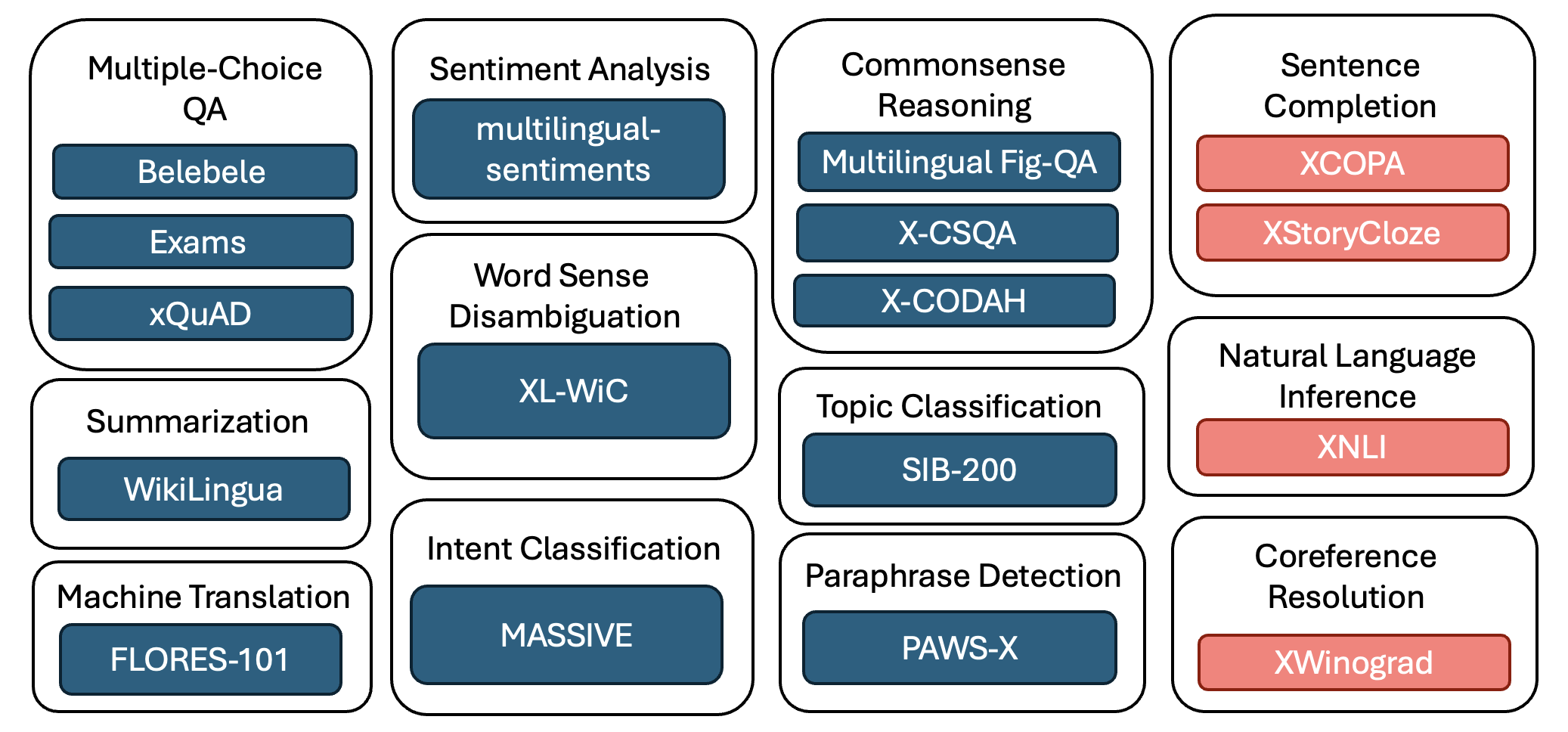}
    \caption{NLU datasets and their respective tasks used for multilingual statement fine-tuning.}
    \label{fig:mult-datasets}
\end{figure}

\section{Experimental Setup}


\subsection{Models}
We experiment with 4 multilingual encoder models of different sizes and multilingual capabilities, namely mBERT \citep{devlin-etal-2019-bert}, XLM-R \citep{conneau-etal-2020-unsupervised}, mDeBERTa-v3 \citep{he2021deberta}, and XLM-V \citep{liang-etal-2023-xlm}. We report the number of parameters and pre-training corpora in Table \ref{tab:encoder-models}.

\renewcommand{\arraystretch}{1.3}
\begin{table}[h!]
\centering
\small
\resizebox{\linewidth}{!}{
\begin{tabular}{ccp{0.35\linewidth}}
\hline
\textbf{Model} & \textbf{\# Params} & \textbf{Pretraining Corpus} \\
\hline
mBERT base & 110M & Wikipedia \\
mDeBERTa-v3 & 276M & \multirow{2}{=}{CC-100 \cite{wenzek-etal-2020-ccnet}}  \\
XLM-R large & 560M &  \\
\hline
\end{tabular}
}
\caption{Multilingual encoder models with their parameter sizes and pretraining corpora.}
\label{tab:encoder-models}
\end{table}

\subsection{Baselines}
\label{sec:baselines}
To assess the capabilities of encoder models with \mst, we compare them with several state-of-the-art instruction-tuned multilingual generative models (except Gemma 2 which is not specifically pretrained for multilingually but has some multilingual support) ranging from 500 million to 72 billion parameters in size. We use models from the following language families: Qwen2 \cite{yang2024qwen2technicalreport}, Llama3.1 \cite{dubey2024llama3herdmodels}, Gemma 2 \cite{gemmateam2024gemma2improvingopen}, and Aya 23 \cite{aryabumi2024aya23openweight}.

We make our comparison through two forms. First, we finetune the base models on an instruction dataset we created using the same datasets and languages used for \mst ~(for details see Appendix~\ref{sec:appendixb}). The second setup involves using the instruction-tuned varieties released by the original team. We include the first setup as a fair comparison controlling for the same fine-tuning data and the second to gauge the performance against the publicly released instruction-tuned versions which employ more data and techniques as another strong baseline for performance.


We make use of 4 unseen multilingual NLU benchmarks in our analysis, XCOPA (commonsense reasoning) \cite{ponti-etal-2020-xcopa}, XNLI (sentence understanding) \cite{conneau-etal-2018-xnli}, XStoryCloze \cite{lin-etal-2022-shot}, and XWinograd \cite{muennighoff-etal-2023-crosslingual, tikhonov2021heads}. Some of the languages in these benchmarks are unseen by our \mst~models, demonstrating cross-lingual generalization. 
For most of the analysis, we report averages across all the languages included in the evaluation datasets, more detailed figures with individual languages are included in Appendix~\ref{sec:lang-level}.

Although we have tried to select models and evaluation data to minimize the chance of leakage of the evaluation and (pre)training data, some (generative) models' pretraining is not completely open, and hence, this remains a limitation of our analysis that is difficult to control. As shown in Table~\ref{tab:encoder-models}, the pretraining of all encoder models is open and, to our knowledge, does not include the evaluation data. For all generative models, we employ the prompting templates provided by the Language Model Evaluation Harness \cite{eval-harness}.

\subsection{Ablations}

As part of our analysis, we ablate several design choices of an encoder-only cross-lingual generalization system. We experiment with encoder models of sizes ranging from 110 million parameters to 560 million (models are outlined in Section~\ref{sft-setup}). 

We explore several design choices for statement generation. First, we use a multilingual prompt template as opposed to just using an English template. To achieve this we machine translate the English template to the language of the example using ChatGPT, specifically the GPT-3.5 version \cite{openai2023chatgpt}. 

Furthermore, we are interested in the effect on cross-lingual generalization when more languages are used during the Statement-tuning step so we explore 3 linguistic setups: English-only (with and without machine translation in the task mixture), 11 languages, and 25 languages to be used during Statement Tuning (languages used for each setup are outlined in Appendix~\ref{language-outline}).

Finally, we directly explore the effect of machine translation data in the \st~training data mixture.
For the rest of the design choices, such as the number of statements to use per dataset and number of templates to use, we follow the general guidelines recommended by \citet{elshabrawy2024enablingnaturalzeroshotprompting}.
Furthermore, we explore the advantages of using encoder models over generative models from an efficiency perspective by exploring the inference time of encoder models against generative models in Section~\ref{sec:time}.

\section{Results and Analysis}
In this section, we derive insights from our experimental results about the cross-lingual zero-shot generalization capabilities of encoder models.

\subsection{Encoder Models are Cross-Task Generalizers}

In Table~\ref{fig:decoder-baselines}, we report the average (over languages) unseen task performance of models trained with \mst~in 11 languages. We contrast this with several instruction-tuned multilingual decoder models, ranging from 500 million to 72 billion parameters, which were instruction-tuned with the same data as our \mst~models. The individual language performance is shown in Appendix~\ref{sec:lang-level}.


The results show that multilingual encoder models are capable of zero-shot cross-task task generalization over a variety of unseen commonsense reasoning and natural language understanding tasks. For XNLI and XStoryCloze, the best-performing encoder models (mDeBERTa and XLM-R Large) outperform most of the generative models examined. More impressively, XLM-R large has an average accuracy of 78.8 on XStoryCloze outperforming the best-performing LLM, Llama3.1 70B, by \textbf{10.5} points despite having $\sim$\textbf{130} times fewer parameters. 

On XNLI the gap is not as large but quite impressively mDeBERTa is the best-performing model at only 276 million parameters outperforming both Qwen2 72B and Llama3.1 70B by around \textbf{5.7} and \textbf{6.1} points on average. For XCOPA, the same best-performing encoder models still maintain impressive results outperforming all the generative models of under 9 billion parameters, and outperforming one of the 70B+ parameter models (Llama3.1 70B). 

In Appendix~\ref{sec:Instruction-Tuned}, we perform the same analysis but on the instruction-tuned varieties of the models released by the teams who trained them. We largely draw the same conclusions with slight variations where certain models on certain tasks perform slightly better/worse. 

\subsection{Encoder Models are Cross-Lingual Generalizers}

When examining individual language performance (see Appendix~\ref{sec:lang-level} for more details) we note that mDeBERTa had less variation \textbf{across} languages in the same task (i.e. there is less disparity between higher and lower resource languages) when compared with generative models. Moreover, mDeBERTa was able to generalize on \textbf{unseen} tasks on languages \textbf{completely unseen} during \mst~if they were seen during pretraining. This further supports the use of state-of-the-art encoder-only models as alternatives to generative models for low-resource languages and cross-lingual generalization on NLU tasks.


Interestingly, mBERT and XLM-R base do not exhibit such performance; at first, it may seem to be an issue of size; however, mDeBERTa has a parameter size similar to XLM-R base, but significantly outperforms it. Hence, we believe that such generalization capabilities require effective pretraining. 

Nevertheless, all encoder models fail to generalize on the XWinograd benchmark, a coreference resolution dataset, achieving mostly random baseline performance. We attribute this to task selection during the \mst~stage, as most of the datasets used may not have sufficient relevance with coreference resolution tasks. The exception might be XL-WiC, which involves word sense disambiguation.

This aligns with the findings of \citet{elshabrawy2024enablingnaturalzeroshotprompting}, who noted that dataset relevance significantly impact a model's ability to generalize effectively.


\begin{figure}
    \centering
    \includegraphics[width=0.9\linewidth]{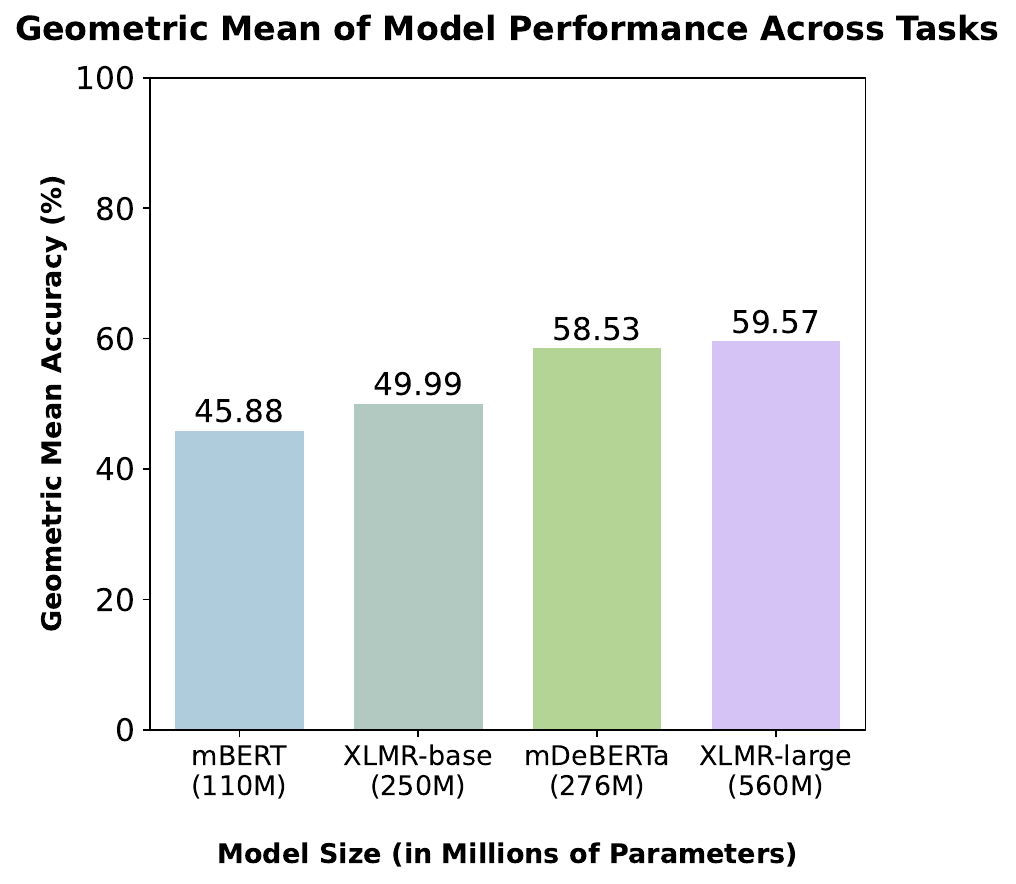}
    \caption{Geometric mean accuracy of multilingual encoder models (mBERT, XLMR-base, mDeBERTa, XLMR-large) across tasks.
    }
    \label{fig:model-ab}
\end{figure}

\subsection{Pretraining and Size Enable Cross-lingual Generalization Abilities}


Interested in generalizing our finding across multiple encoder-only models, we train 4 different models (mBERT, XLMR-base, mDeBERTa, XLMR-large) using \mst~ using the exact same data setup and varying certain hyperparamters depending on model (see Appendix~\ref{sec:appendixb}) until convergence. We evaluate them on the same four unseen benchmark datasets outlined in Section~\ref{sec:baselines}.

In Figure~\ref{fig:model-ab}, we compare the geometric mean of the task performance of the four multilingual encoder-only models we examine with~\st, as discussed earlier we note that the two models mDeBERTa and XLM-R Large exhibit much higher task performance than the other two models mBERT and XLM-R base. Despite finetuning all the models until convergence and performing hyperparameter optimization with all models, this remains the case. 
\begin{figure}
    \centering
    \includegraphics[width=0.75\linewidth]{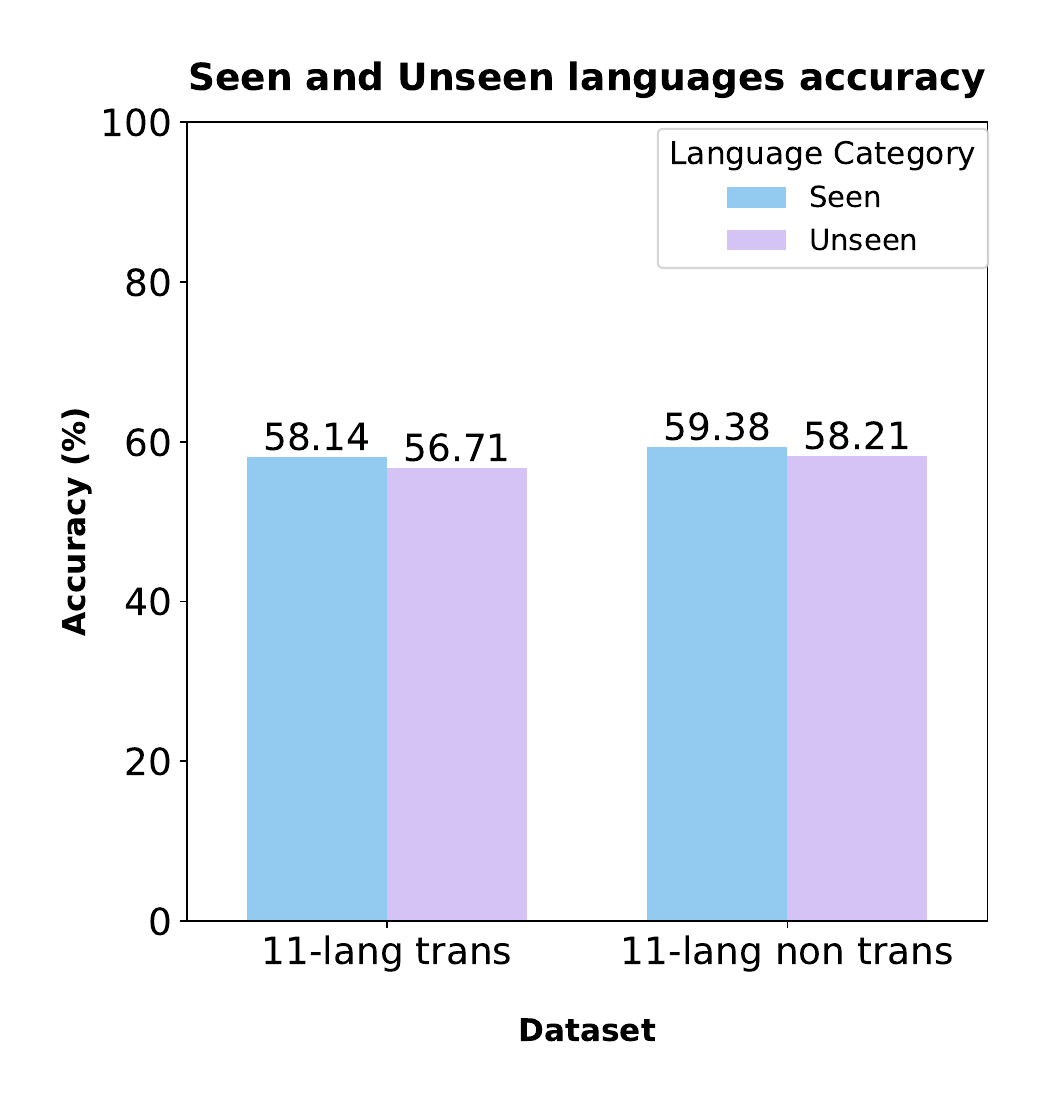}
    \caption{Mean accuracy of mDeBERTa on seen vs. unseen languages during \st~across tasks and languages. 11-lang trans uses machine-translated prompt templates while 11-lang non trans shows performance using English-only prompt templates.}
    \label{fig:seenvunseen}
\end{figure}
Previous work has shown the relatively limited capabilities of mBERT in comparison to other models \cite{conneau-etal-2020-unsupervised} which has different pretraining data and regimes. However, the difference in abilities between XLM-R base and XLM-R large cannot be attributed to just pretraining, as XLM-R base fails to achieve zero-shot cross-lingual generalization with the same pretraining choices as XLM-R large at a different scale. 

It is also difficult to attribute cross-lingual generalization capabilities purely to model size as mDeBERTa achieves similar performance and cross-lingual generalization capabilities as XLM-R large but at a size comparable to XLM-R base. Hence, we hypothesize that cross-lingual zero-shot generalization (being an inherently difficult task) emerges in encoder-only as a function of both size and pretraining. In general, encoder models that have shown state-of-the-art performance on general tasks are more likely to exhibit cross-lingual generalization capabilities but it is not strictly a matter of model "capacity" as would be implied by model size, or pretraining data.

\subsection{English-only Prompting Templates are Sufficient to Enable Effective \st}

\begin{figure}
    \centering
    \includegraphics[width=0.9\linewidth]{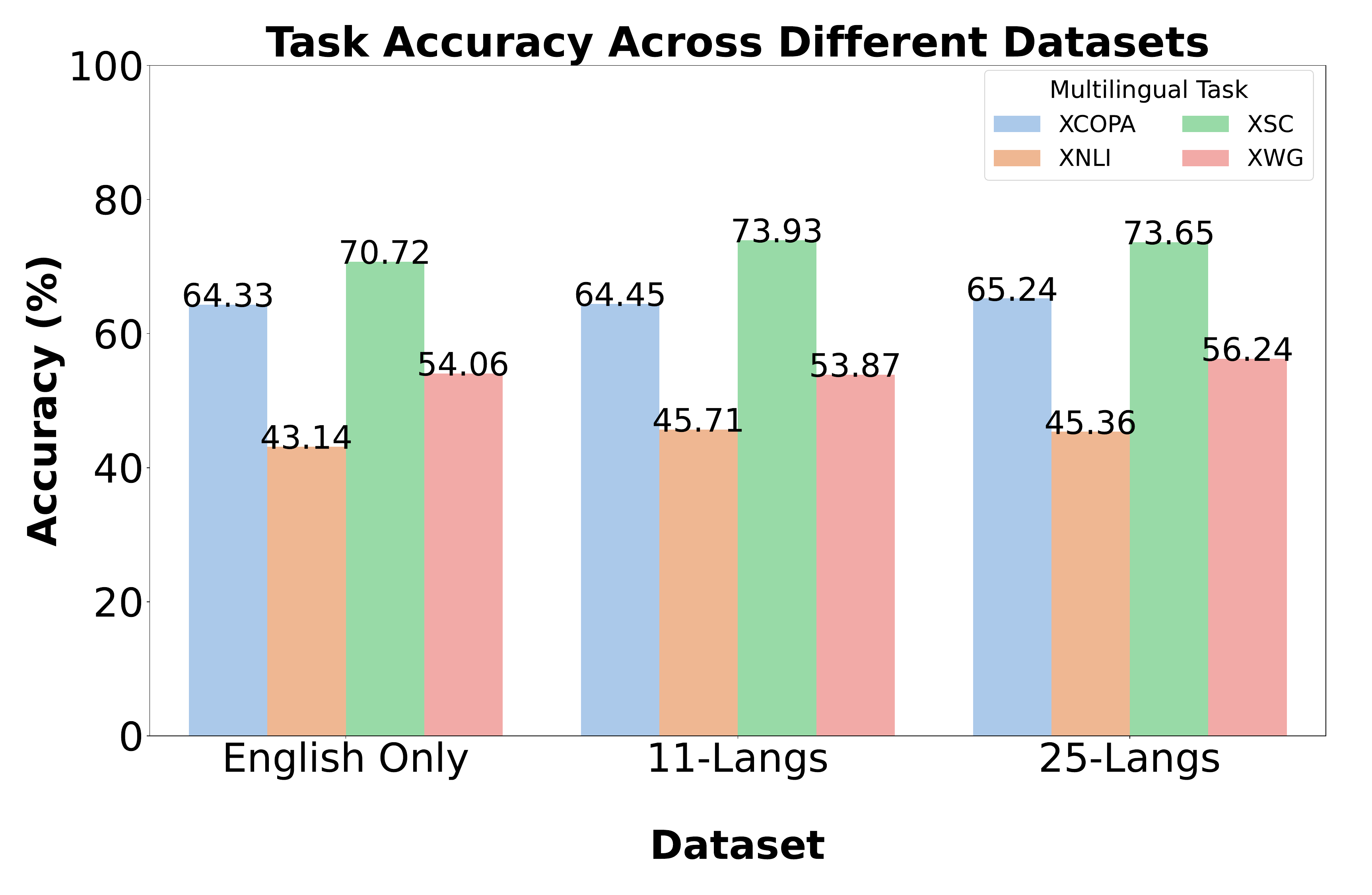}
    \caption{Task accuracy across different training datasets (English-only, 11-langs, and 25-langs) using mDeBERTa.
    }
    \label{fig:lang-ablation}
\end{figure}
As part of our investigation on assessing the cross-lingual zero-shot generalization capabilities of encoder-only models, we experiment to see if the use of machine-translated prompt templates would offer any improved performance over the use of English-only prompt templates for the various tasks. 
To achieve this, we utilized ChatGPT, specifically the GPT-3.5 version \cite{openai2023chatgpt}, to machine-translate each prompt to match the language of the example being turned into a statement. We then train and evaluate a model using these machine-translated examples to assess the impact on performance across different languages. An obvious limitation of this analysis could be the quality of the Machine Translation, however, we decided to use a commonly accessible model.

\begin{figure}
    \centering
    \includegraphics[width=0.85\linewidth]{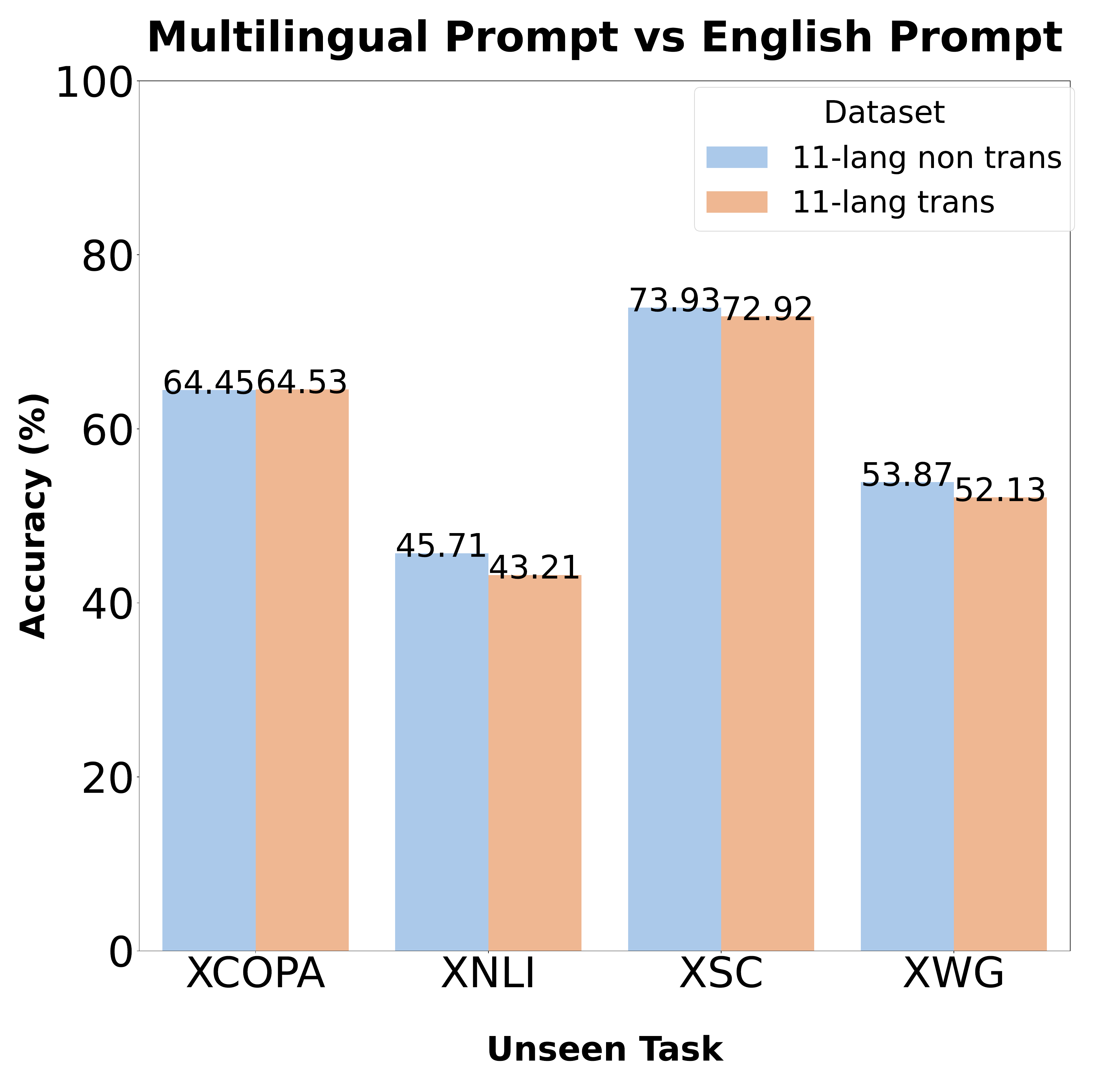}
    \caption{Mean task accuracy of mDeBERTa (finetuned on 11 languages during \st) over languages on the 4 evaluation tasks using machine translated prompt templates (11-lang trans) vs. English-only prompt templates (11-lang non trans).}
    \label{fig:translate-prompt}
\end{figure}

As seen in Figure~\ref{fig:translate-prompt}, we do not observe any apparent benefit to translating prompts. This is consistent with observations from multilingual prompting of LLMs \cite{zhang2023promptinglargelanguagemodel}. 

Curious to see if perhaps machine translating prompt templates helps benefit generalization capabilities we compare seen versus unseen (during \st) average language performance in Figure~\ref{fig:seenvunseen}. Again, we observe no apparent benefit of unseen languages. This observation has the added benefit of simplifying prompt design and reducing the computational/time cost of having to machine translate prompts.

\subsection{Multilingual Pretraining Sufficiently Enables Cross-lingual Generalization}
\label{sec:language-setups}
To understand the effect of Multilinguality during \st~on cross-lingual zero-shot task generalization we experiment by changing the number of languages included in the \st~ data by using the same training tasks with a differing number of languages being included in the training set.

In Figure~\ref{fig:lang-ablation}, we examine the effect of including more languages in the training set on cross-lingual capabilities. We examine 3 setups, English-only where the mDeBERTa model is only trained on the English subsets/equivalents of the training datasets (except for the machine translation task which \textbf{is} included), 11-langs where the model is trained on subsets of the data that belong to only 11 of the possible 25 languages and 25-langs which includes all possible languages in the training datasets. By sampling, we fix for training set sizes to be similar in size with the English-only and 11-langs to include 123k training examples and the 25-langs dataset including $\sim$185k examples (it needed to be slightly larger to representatively sample the languages). 

Overall, we observe that most of the cross-lingual task performance can actually be obtained by finetuning a multilingual encoder model on a single language (English) multi-task statement dataset, with the English-only setup achieving \textbf{98.6\%}, \textbf{95.1\%}, and \textbf{96.0\%} of the performance of 25-langs on XCOPA, XNLI, and XStoryCloze respectively. Increasing the number of represented languages to 11 during \st~ yields gains over English-only and manages to very slightly outperform using all languages on XNLI and XStoryCloze. 

Furthermore, in Figure~\ref{fig:seenvunseen}, we compare the average performance of the 11-lang model on seen versus unseen languages during \st. We only observe a slight performance gain on average (59.4 versus 58.2) when languages are seen during \st. This leads us to believe that most of the cross-lingual generalization capabilities are impressively due to the multilingual pre-training, rather than requiring cross-lingual exposure during \st. This opens up many potential use cases of encoder-only models for zero-shot task generalization for use with languages without necessitating any supervised \st~ in these languages which proves very useful given the abundance of task data in high-resource languages such as English.

Nevertheless, we report the performance of 11-langs on other experimental setups as an intermediate between both extremes while also allowing us to observe differences in performance between seen/unseen languages during~\st.

\subsection{Including Machine Translation in \st~Training Data improves Cross-lingual Transfer}

\begin{figure}
    \centering
    \includegraphics[width=\linewidth]{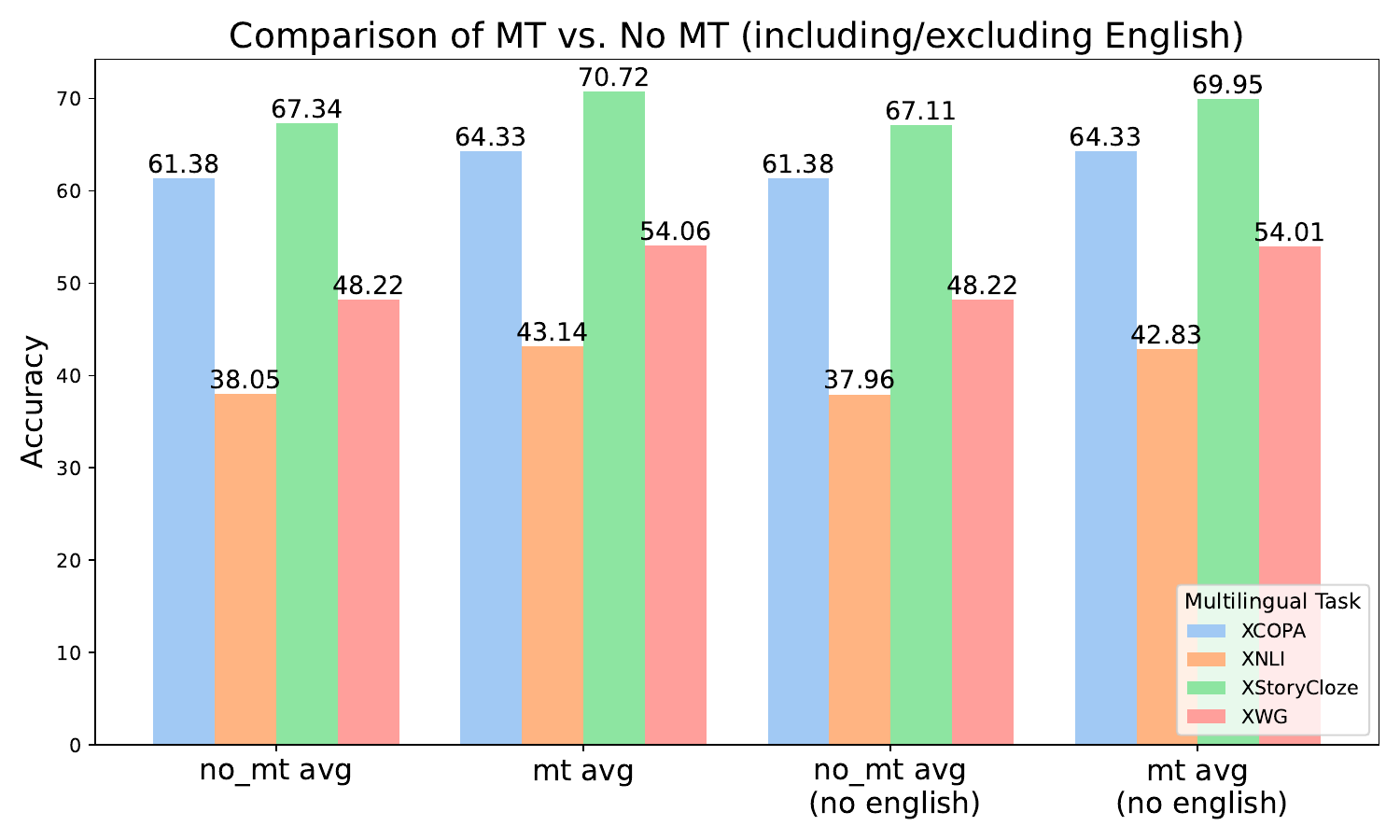}
    \caption{Mean task accuracy of mDeBERTa on an English-only task dataset (including and excluding MT statement data). On the left are the averages including English evaluation sets and on the right excluding them.}
    \label{fig:mt-no_mt}
\end{figure}

In section~\ref{sec:language-setups}, we observed that an English-only training setup that includes machine translation (MT) data can achieve most of the performance benefits of using multiple languages in the training task mixture. However, it remains unclear whether the inclusion of MT data itself is a key contributing factor. To investigate this, Figure~\ref{fig:mt-no_mt} directly compares the effect of including versus excluding MT data in the English-only setup.

Interestingly, incorporating MT data leads to a notable performance increase across all tasks, both in the average performance across all languages and in the average performance excluding English (the "seen" language) from the evaluation. This suggests that MT data enhances cross-lingual transfer and is particularly beneficial when language-specific NLU task data is unavailable. In such cases, using English-only task data with MT achieves a significant portion of the multilingual performance. Additionally, since MT data is generally more accessible for lower-resource languages than NLU task data, it is relatively easy to incorporate into the \st~training mix.

\subsection{Multilingual \st~Enables Efficient Inference for Zero-shot Cross-lingual Generalization}
\label{sec:time}
Though \st~ enables generalization into zero-shot settings with comparable performance against the zero-shot LLMs, increasing the number of candidate labels would also increase the model's computational overhead. In the case of a statement-tuned model, for a downstream task with $n$-possible labels, a naive way to perform a prediction is to iterate the model over $n$-times for each label. But in practice, it'd be more efficient to perform a batched prediction. Here, we compare the inference time and maximum batch sizes of our best statement-tuned model (mDeBERTa-v3-base; 0.276B) against zero-shot LLMs with varying parameter sizes, ranging from 0.5B to 9B. 

\begin{figure}
    \centering
    \includegraphics[width=0.95\linewidth]{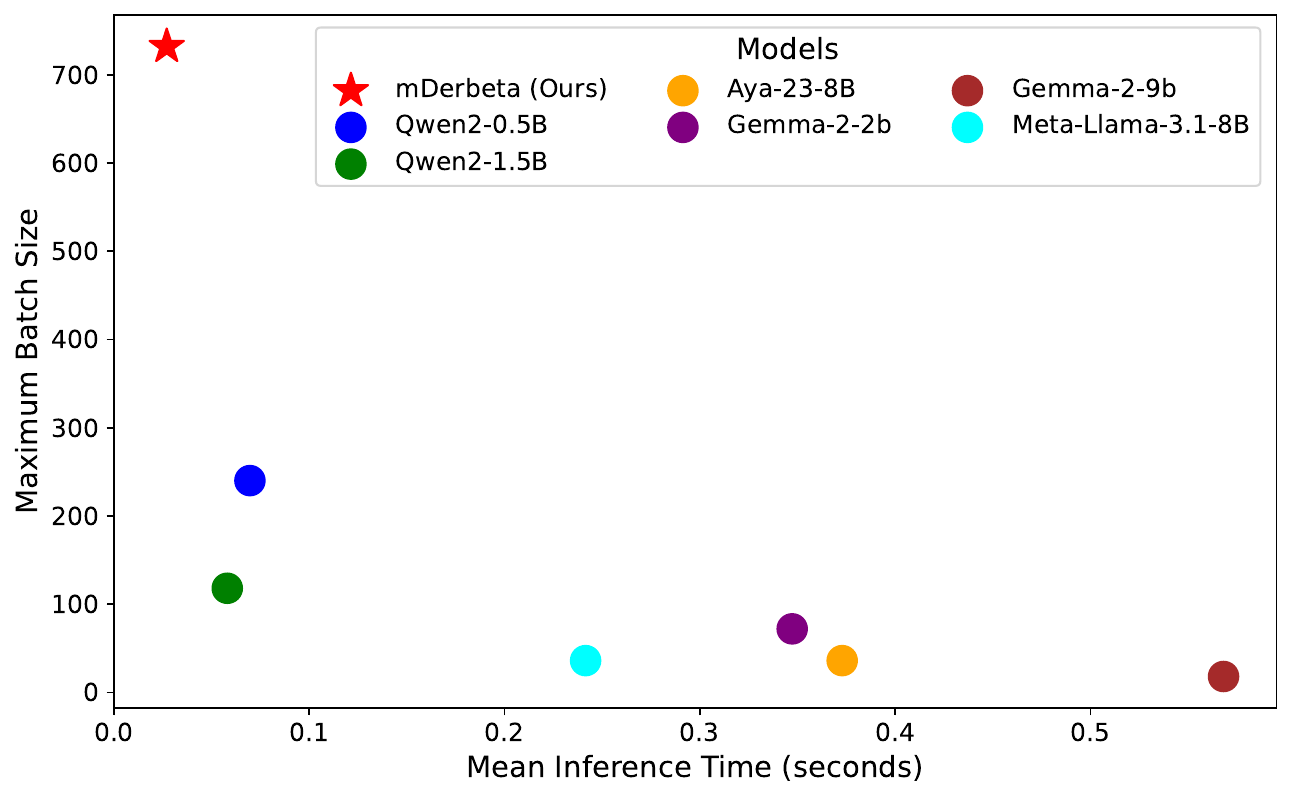}
    \caption{Mean inference time and maximum batch size of various models during a simulated text classification task on a single A100 GPU.}
    \label{fig:inference-time}
\end{figure}

We simulate a text classification task on each model, measuring the mean inference time per batch where we gradually increase the batch size. We perform this experiment on a single A100 GPU and show the result in Figure~\ref{fig:inference-time}. As expected, due to our model size and non-autoregressive nature, it achieves the fastest mean inference time with the largest batch size capacity. Having a statement-tuned model that could handle $m$-batch size means that it could handle $m / n$ instances at once.



\section{Conclusion}
While large generative models dominate multilingual NLP, the potential of encoder-only models for cross-lingual generalization remains underexplored. We show that a well-designed finetuning setup enables state-of-the-art pretrained encoder-only models to match, or even surpass, generative models in three of four unseen cross-lingual NLU tasks, despite using far fewer parameters. Additionally, these models generalize across languages even when finetuned only on a monolingual multi-task dataset, leveraging their multilingual pretraining. Our findings position encoder-only models as a memory-efficient alternative for multilingual multitask NLU. Future work can further optimize finetuning, extend cross-lingual generalization, and refine encoder architectures for large-scale multilingual learning.


\section*{Limitations}
\st~highly relies on the training task selection and the proximity of the chosen tasks to the target task, hence the utility of the approach may still be limited if the training task selection fails to include similar enough examples to the target task. Due to this, some tasks like XWinograd may not be sufficiently addressed.

\st~requires the use of verbalization which requires extra effort and careful prompt design. Furthermore, requiring a statement for each potential target makes this method infeasible for tasks with an extremely large hypothesis class. 

Not all the encoder models we studied were capable of cross-lingual generalization and we were not able to pinpoint the exact mechanism during pretraining which enables such capabilities. We leave this for future work.

We were not able to control for the pretraining/instruction-tuning of all the models explored due to the lack of transparency regarding exact training data for some models. Hence, our analysis may include models which are not completely blind to the task data.

\bibliography{anthology,custom}
\bibliographystyle{acl_natbib}
\clearpage

\appendix
\section{Statement Templates}
\label{sec:appendixa}
\subsection{Multiple-Choice QA Templates}
\subsubsection{Belebele Templates}

\renewcommand{\arraystretch}{}
\resizebox{\columnwidth}{!}{
    \begin{tabular}{ll}\toprule
    \textbf{Task} &\textbf{Statement Template} \\\midrule
    \multirow{24}{*}{Belebele} & \{\{context\}\} \{\{question\}\} \{\{correct\_answer/other\_answer\}\} \\
    & \{\{context\}\} According to the passage above, the answer of \{\{question\}\} is \{\{correct\_answer/other\_answer\}\} \\
    & Passage: \{\{context\}\} Question: \{\{question\}\} Answer: \{\{correct\_answer/other\_answer\}\} \\
    & \{\{context\}\} Q: \{\{question\}\} A: \{\{correct\_answer/other\_answer\}\} \\

    & Content: "\{\{context\}\}"\textbackslash n Inquiry: "\{\{question\}\}"\textbackslash n Response: "\{\{correct\_answer/other\_answer\}\}" \\
    & Text: "\{\{context\}\}"\textbackslash n Query: "\{\{question\}\}"\textbackslash n Solution: "\{\{correct\_answer/other\_answer\}\}" \\
    & Passage content: "\{\{context\}\}"\textbackslash n What is asked: "\{\{question\}\}"\textbackslash n The answer is: "\{\{correct\_answer/other\_answer\}\}" \\
    & Here is the passage: "\{\{context\}\}"\textbackslash n The question is: "\{\{question\}\}"\textbackslash n The provided answer is: "\{\{correct\_answer/other\_answer\}\}" \\
    & The passage reads: "\{\{context\}\}"\textbackslash n Asked: "\{\{question\}\}"\textbackslash n The correct answer is: "\{\{correct\_answer/other\_answer\}\}" \\
    & From the text: \{\{context\}\}\textbackslash n As stated above, the response to "\{\{question\}\}" is "\{\{correct\_answer/other\_answer\}\}" \\
    & Based on the passage: \{\{context\}\}\textbackslash n The answer to "\{\{question\}\}" according to the text is "\{\{correct\_answer/other\_answer\}\}" \\
    & The content: \{\{context\}\}\textbackslash n Thus, the answer to "\{\{question\}\}" is "\{\{correct\_answer/other\_answer\}\}" \\
    & In reference to the passage: \{\{context\}\}\textbackslash n According to the text, the answer for "\{\{question\}\}" is "\{\{correct\_answer/other\_answer\}\}" \\
    & Given the text: \{\{context\}\}\textbackslash n Therefore, the answer to "\{\{question\}\}" is "\{\{correct\_answer/other\_answer\}\}" \\
    & Text: \{\{context\}\}\textbackslash n Inquiry: \{\{question\}\}\textbackslash n Response: \{\{correct\_answer/other\_answer\}\} \\
    & Content: \{\{context\}\}\textbackslash n Question asked: \{\{question\}\}\textbackslash n Given Answer: \{\{correct\_answer/other\_answer\}\} \\
    & Passage: \{\{context\}\}\textbackslash n Question: \{\{question\}\}\textbackslash n Answer Provided: \{\{correct\_answer/other\_answer\}\} \\
    & The text reads: \{\{context\}\}\textbackslash n Query: \{\{question\}\}\textbackslash n Solution: \{\{correct\_answer/other\_answer\}\} \\
    & Content: \{\{context\}\}\textbackslash n What is the question: \{\{question\}\}\textbackslash n The answer is: \{\{correct\_answer/other\_answer\}\} \\
    & \{\{context\}\}\textbackslash n Question: \{\{question\}\}\textbackslash n Answer: \{\{correct\_answer/other\_answer\}\} \\
    & \{\{context\}\}\textbackslash n Inquiry: \{\{question\}\}\textbackslash n Response: \{\{correct\_answer/other\_answer\}\} \\
    & \{\{context\}\}\textbackslash n What is being asked: \{\{question\}\}\textbackslash n The answer is: \{\{correct\_answer/other\_answer\}\} \\
    & \{\{context\}\}\textbackslash n The question posed is: \{\{question\}\}\textbackslash n The correct answer is: \{\{correct\_answer/other\_answer\}\} \\
    & \{\{context\}\}\textbackslash n The text asks: \{\{question\}\}\textbackslash n The response provided is: \{\{correct\_answer/other\_answer\}\} \\
    \bottomrule
    \end{tabular}
}

\subsubsection{Exams Templates}

\resizebox{\columnwidth}{!}{
\begin{tabular}{ll}\toprule
\textbf{Task} &\textbf{Statement Template} \\\midrule
    \multirow{3}{*}{Exams} & Q: \{\{question\}\}. A: \{\{correct\_answer/other\_answer\}\} \\
    & \{\{question\}\}. Answer: \{\{correct\_answer/other\_answer\}\} \\
    & Question: \{\{question\}\} Answer: \{\{correct\_answer/other\_answer\}\} \\
\bottomrule
\end{tabular}
}

\subsubsection{xQuAD Templates}

\resizebox{\columnwidth}{!}{
\begin{tabular}{ll}\toprule
\textbf{Task} &\textbf{Statement Template} \\\midrule
    \multirow{20}{*}{xQuAD} & \{\{context\}\} Question: \{\{question\}\} Answer: \{\{correct\_answer/other\_answer\}\} \\
    & Passage: \{\{context\}\} Question: \{\{question\}\} Answer: \{\{correct\_answer/other\_answer\}\} \\
    & \{\{context\}\} Q: \{\{question\}\} A: \{\{correct\_answer/other\_answer\}\} \\
    & \{\{context\}\} According to the passage above, the answer of \{\{question\}\} is \{\{correct\_answer/other\_answer\}\} \\
    & Text: \{\{context\}\}\textbackslash n Question: \{\{question\}\}\textbackslash n Reply: \{\{correct\_answer/other\_answer\}\} \\
    & Passage text: \{\{context\}\}\textbackslash n What is the solution: \{\{question\}\}\textbackslash n Answer: \{\{correct\_answer/other\_answer\}\} \\
    & \{\{context\}\}\textbackslash n In reference to the text, what is the answer: \{\{question\}\}\textbackslash n Answer: \{\{correct\_answer/other\_answer\}\} \\
    & From the given passage: \{\{context\}\}\textbackslash n Query: \{\{question\}\}\textbackslash n Solution: \{\{correct\_answer/other\_answer\}\} \\
    & Passage: \{\{context\}\}\textbackslash n Q: \{\{question\}\}\textbackslash n A: \{\{correct\_answer/other\_answer\}\} \\
    & Text: \{\{context\}\}\textbackslash n What is the response: \{\{question\}\}\textbackslash n Answer: \{\{correct\_answer/other\_answer\}\} \\
    & Context: \{\{context\}\}\textbackslash n Answer for \{\{question\}\} is: \{\{correct\_answer/other\_answer\}\} \\
    & According to the context: \{\{context\}\}\textbackslash n Solution to \{\{question\}\}: \{\{correct\_answer/other\_answer\}\} \\
    & Text: \{\{context\}\}\textbackslash n Answer to \{\{question\}\} is: \{\{correct\_answer/other\_answer\}\} \\
    & \{\{context\}\}\textbackslash n In reference to the passage, \{\{question\}\} has the answer: \{\{correct\_answer/other\_answer\}\} \\
    & \{\{context\}\}\textbackslash n Answer to the question \{\{question\}\} based on the passage is: \{\{correct\_answer/other\_answer\}\} \\
    & \{\{context\}\}\textbackslash n From the passage, the response to \{\{question\}\} is: \{\{correct\_answer/other\_answer\}\} \\
    & Text: \{\{context\}\}\textbackslash n Question: \{\{question\}\}\textbackslash n Response: \{\{correct\_answer/other\_answer\}\} \\
    & Passage: \{\{context\}\}\textbackslash n Query: \{\{question\}\}\textbackslash n Answer: \{\{correct\_answer/other\_answer\}\} \\
    & Context: \{\{context\}\}\textbackslash n What is the answer to \{\{question\}\}?\textbackslash n Solution: \{\{correct\_answer/other\_answer\}\} \\
    & From the text: \{\{context\}\}\textbackslash n What is the solution to \{\{question\}\}?\textbackslash n Answer: \{\{correct\_answer/other\_answer\}\} \\
\bottomrule
\end{tabular}
}

\subsection{Summarization Templates}
\subsubsection{WikiLingua Templates}

\resizebox{\columnwidth}{!}{
\begin{tabular}{ll}\toprule
\textbf{Task} &\textbf{Statement Template} \\\midrule
    \multirow{4}{*}{WikiLingua} & Passage: \{\{source\}\}, Summary: \{\{correct\_target/random\_target\}\} \\
    & The summary of "\{\{source\}\}" is \{\{correct\_target/random\_target\}\} \\
    & Context: \{\{source\}\}, Summary: \{\{correct\_target/random\_target\}\} \\
    & Q: Summarize the following: \{\{source\}\}, A: \{\{correct\_target/random\_target\}\} \\
    & The answer of "Summarize the following \{\{source\}\}" is \{\{correct\_target/random\_target\}\} \\
\bottomrule
\end{tabular}
}

\subsection{Machine Translation Templates}
\subsubsection{FLORES-101 Templates}

\resizebox{\columnwidth}{!}{
\begin{tabular}{ll}\toprule
\textbf{Task} &\textbf{Statement Template} \\\midrule
    \multirow{2}{*}{FLORES-101} & The \{\{target\_lang\}\} translation of \{\{lang\}\} sentence \{\{sentence\}\} is \{\{target\_sentence\}\} \\
    & The \{\{target\_lang\}\} translation of \{\{lang\}\} sentence \{\{sentence\}\} is not \{\{target\_sentence\}\} \\

\bottomrule
\end{tabular}
}

\subsection{Sentiment Analysis Templates}
\subsubsection{multilingual-sentiments Templates}

\resizebox{\columnwidth}{!}{
\begin{tabular}{ll}\toprule
\textbf{Task} &\textbf{Statement Template} \\\midrule
\multirow{10}{*}{multilingual-sentiments} & The text '\{\{text\}\}' is \{\{correct\_label/other\_label\}\}. \\
    & Sentence: '\{\{text\}\}'. Label: \{\{correct\_label/other\_label\}\} \\
    & Sentiment Analysis:\textbackslash nText: \{\{text\}\}\textbackslash nResult: \{\{correct\_label/other\_label\}\} \\
    & The sentiment of the text \{\{text\}\} is \{\{correct\_label/other\_label\}\} \\
    & Text: \{\{text\}\} has a sentiment labeled as \{\{correct\_label/other\_label\}\} \\
    & The text \{\{text\}\} conveys a sentiment of \{\{correct\_label/other\_label\}\} \\
    & The analysis reveals that \{\{text\}\} is characterized by a sentiment of \{\{correct\_label/other\_label\}\} \\
    & For the text \{\{text\}\}, the sentiment is identified as \{\{correct\_label/other\_label\}\} \\
    & The sentiment associated with \{\{text\}\} is \{\{correct\_label/other\_label\}\} \\
    & In terms of sentiment, \{\{text\}\} reflects \{\{correct\_label/other\_label\}\} \\
\bottomrule
\end{tabular}
}

\subsection{Word Sense Disambiguation Templates}
\subsubsection{XL-WiC Templates}

\resizebox{\columnwidth}{!}{
\begin{tabular}{ll}\toprule
\textbf{Task} &\textbf{Statement Template} \\\midrule
    \multirow{12}{*}{XL-WiC} & "\{\{target\_word\}\}" means the same in "\{\{context\_1\}\}" and "\{\{context\_2\}\}" \\
    & "\{\{target\_word\}\}" does not mean the same in "\{\{context\_1\}\}" and "\{\{context\_2\}\}" \\
    & The meaning of "\{\{target\_word\}\}" is consistent across "\{\{context\_1\}\}" and "\{\{context\_2\}\}" \\
    & The meaning of "\{\{target\_word\}\}" is inconsistent across "\{\{context\_1\}\}" and "\{\{context\_2\}\}" \\
    & The interpretation of "\{\{target\_word\}\}" remains unchanged in both "\{\{context\_1\}\}" and "\{\{context\_2\}\}" \\
    & The interpretation of "\{\{target\_word\}\}" changes in "\{\{context\_1\}\}" and "\{\{context\_2\}\}" \\
    & The sense of \{\{target\_word\}\} is identical between \{\{context\_1\}\} and \{\{context\_2\}\} \\
    & The sense of \{\{target\_word\}\} differs between \{\{context\_1\}\} and \{\{context\_2\}\} \\
    & The interpretation of \{\{target\_word\}\} is the same in both \{\{context\_1\}\} and \{\{context\_2\}\} \\
    & The sense of \{\{target\_word\}\} varies between \{\{context\_1\}\} and \{\{context\_2\}\} \\
    & \{\{target\_word\}\} has the same meaning in both \{\{context\_1\}\} and \{\{context\_2\}\} \\
    & The meaning of \{\{target\_word\}\} is different in \{\{context\_1\}\} and \{\{context\_2\}\} \\
\bottomrule
\end{tabular}
}

\subsection{Intent Classification Templates}
\subsubsection{MASSIVE Templates}

\resizebox{\columnwidth}{!}{
\begin{tabular}{ll}\toprule
\textbf{Task} &\textbf{Statement Template} \\\midrule
    \multirow{4}{*}{MASSIVE} & The utterance "\{\{utt\}\}" is under the "\{\{scenario\}\}" scenario. \\
    & Utterance: "\{\{utt\}\}" Scenario: "\{\{scenario\}\}" \\
    & User: "\{\{utt\}\}". The best scenario for the user query is "\{\{scenario\}\}". \\
    & The scenario of user’s utterance "\{\{utt\}\}" is "\{\{scenario\}\}". \\

\bottomrule
\end{tabular}
}

\subsection{Commonsense Reasoning Templates}
\subsubsection{Multilingual Fig-QA Templates}

\resizebox{\columnwidth}{!}{
\begin{tabular}{ll}\toprule
\textbf{Task} &\textbf{Statement Template} \\\midrule
    \multirow{6}{*}{Multilingual Fig-QA} & "\{\{startphrase\}\}" "\{\{ending1/ending2\}\}" \\
    & "\{\{startphrase\}\}" therefore "\{\{ending1/ending2\}\}" \\
    & Startphrase: "\{\{startphrase\}\}" ending: "\{\{ending1/ending2\}\}" \\
    & "\{\{startphrase\}\}" then "\{\{ending1/ending2\}\}" \\
    & if "\{\{startphrase\}\}" then "\{\{ending1/ending2\}\}" \\
    & "\{\{startphrase\}\}" means "\{\{ending1/ending2\}\}" \\
\bottomrule
\end{tabular}
}

\subsubsection{X-CSQA Templates}

\resizebox{\columnwidth}{!}{
\begin{tabular}{ll}\toprule
\textbf{Task} &\textbf{Statement Template} \\\midrule
    \multirow{15}{*}{X-CSQA} & Question: "\{\{question\}\}". Answer: "\{\{correct\_answer/other\_answer\}\}" \\
    & Q: "\{\{question\}\}". A: "\{\{correct\_answer/other\_answer\}\}" \\
    & "\{\{question\}\}". Ans: "\{\{correct\_answer/other\_answer\}\}" \\
    & Inquiry: \{\{question\}\}\textbackslash nResponse: "\{\{correct\_answer/other\_answer\}\}" \\
    & The question: \{\{question\}\} has the answer: "\{\{correct\_answer/other\_answer\}\}" \\
    & Question posed: "\{\{question\}\}". Possible response: "\{\{correct\_answer/other\_answer\}\}" \\
    & In response to \{\{question\}\}, the answer is "\{\{correct\_answer/other\_answer\}\}" \\
    & Query: \{\{question\}\}\textbackslash nResponse: "\{\{correct\_answer/other\_answer\}\}" \\
    & The query: \{\{question\}\} yields the answer: "\{\{correct\_answer/other\_answer\}\}" \\
    & The answer to \{\{question\}\} could be: "\{\{correct\_answer/other\_answer\}\}" \\
    & For the question: \{\{question\}\}, the answer is "\{\{correct\_answer/other\_answer\}\}" \\
    & The inquiry \{\{question\}\} could receive the answer: "\{\{correct\_answer/other\_answer\}\}" \\
    & The query: \{\{question\}\} has the answer: "\{\{correct\_answer/other\_answer\}\}" \\
    & When posed with the question: \{\{question\}\}, the answer provided is "\{\{correct\_answer/other\_answer\}\}" \\
    & Upon inquiry: \{\{question\}\}, the answer provided is "\{\{correct\_answer/other\_answer\}\}" \\
\bottomrule
\end{tabular}
}

\subsubsection{X-CODAH Templates}

\resizebox{\columnwidth}{!}{
\begin{tabular}{ll}\toprule
\textbf{Task} &\textbf{Statement Template} \\\midrule
    \multirow{18}{*}{X-CODAH} & The statement \{\{correct\_text\}\} makes more sense than the statement \{\{other\_text\}\}. \\
    & Statement \{\{correct\_text\}\} is more logical than \{\{other\_text\}\}. \\
    & The statement \{\{correct\_text\}\} makes sense. \\
    & The statement \{\{correct\_text\}\} is clearer compared to \{\{other\_text\}\}. \\
    & \{\{correct\_text\}\} is more reasonable than \{\{other\_text\}\}. \\
    & Between the two, \{\{correct\_text\}\} is the more sensible statement over \{\{other\_text\}\}. \\
    & \{\{correct\_text\}\} presents a clearer rationale than \{\{other\_text\}\}. \\
    & When comparing, \{\{correct\_text\}\} is more coherent than \{\{other\_text\}\}. \\
    & Statement \{\{correct\_text\}\} exhibits greater logic than \{\{other\_text\}\}. \\
    & In terms of logic, \{\{correct\_text\}\} surpasses \{\{other\_text\}\}. \\
    & \{\{correct\_text\}\} shows a higher degree of logical reasoning than \{\{other\_text\}\}. \\
    & Compared to \{\{other\_text\}\}, statement \{\{correct\_text\}\} is the more logical choice. \\
    & Between the two, \{\{correct\_text\}\} is the more logical statement compared to \{\{other\_text\}\}. \\
    & The statement \{\{correct\_text\}\} is sensible and coherent. \\
    & Clearly, the statement \{\{correct\_text\}\} is logical. \\
    & It is evident that the statement \{\{correct\_text\}\} is reasonable. \\
    & Undoubtedly, the statement \{\{correct\_text\}\} holds logic. \\
    & There is clarity in the statement \{\{correct\_text\}\}. \\
\bottomrule
\end{tabular}
}

\subsection{Topic Classification Templates}
\subsubsection{SIB-200 Templates}

\resizebox{\columnwidth}{!}{
\begin{tabular}{ll}\toprule
\textbf{Task} &\textbf{Statement Template} \\\midrule
\multirow{42}{*}{SIB-200} & Sentence: \{\{text\}\}. Label: \{\{label\}\}. \\
    & The sentence \{\{text\}\} is considered a \{\{label\}\} sentence. \\
    & The sentence \{\{text\}\} is not considered a \{\{label\}\} sentence. \\
    & The sentence \{\{text\}\} is about \{\{label\}\}. \\
    & The sentence \{\{text\}\} is not about \{\{label\}\}. \\
    & The sentence \{\{text\}\} is a \{\{label\}\} sentence. \\
    & The sentence \{\{text\}\} is not a \{\{label\}\} sentence. \\
    & Text: \{\{text\}\} \textbackslash n Category: \{\{label\}\}. \\
    & The text: "\{\{text\}\}" is labeled as \{\{label\}\}. \\
    & Sentence: "\{\{text\}\}" \textbackslash n Topic: \{\{label\}\}. \\
    & The given sentence "\{\{text\}\}" belongs to the category: \{\{label\}\}. \\
    & The sentence describes a \{\{label\}\} topic: \{\{text\}\}. \\
    & The text "\{\{text\}\}" discusses \{\{label\}\}. \\
    & "\{\{text\}\}" talks about the topic: \{\{label\}\}. \\
    & This sentence, "\{\{text\}\}", revolves around \{\{label\}\}. \\
    & "\{\{text\}\}" is centered on \{\{label\}\}. \\
    & The topic of "\{\{text\}\}" is \{\{label\}\}. \\
    & The text "\{\{text\}\}" does not discuss \{\{label\}\}. \\
    & "\{\{text\}\}" does not talk about \{\{label\}\}. \\
    & This sentence, "\{\{text\}\}", does not revolve around \{\{label\}\}. \\
    & "\{\{text\}\}" is not related to \{\{label\}\}. \\
    & The topic of "\{\{text\}\}" is not \{\{label\}\}. \\
    & The text "\{\{text\}\}" is regarded as a \{\{label\}\} sentence. \\
    & "\{\{text\}\}" is classified as a \{\{label\}\} sentence. \\
    & This sentence, "\{\{text\}\}", is viewed as \{\{label\}\}. \\
    & The text is recognized as \{\{label\}\}: "\{\{text\}\}". \\
    & The classification of "\{\{text\}\}" is \{\{label\}\}. \\
    & The text "\{\{text\}\}" is not regarded as a \{\{label\}\} sentence. \\
    & "\{\{text\}\}" is not classified as a \{\{label\}\} sentence. \\
    & This sentence, "\{\{text\}\}", is not viewed as \{\{label\}\}. \\
    & The text is not recognized as \{\{label\}\}: "\{\{text\}\}". \\
    & The classification of "\{\{text\}\}" is not \{\{label\}\}. \\
    & The sentence "\{\{text\}\}" falls under the category of \{\{label\}\}. \\
    & "\{\{text\}\}" is labeled as a \{\{label\}\} sentence. \\
    & This sentence, "\{\{text\}\}", is classified as \{\{label\}\}. \\
    & The text "\{\{text\}\}" belongs to the \{\{label\}\} category. \\
    & "\{\{text\}\}" is a sentence of the \{\{label\}\} type. \\
    & The sentence "\{\{text\}\}" does not fall under the category of \{\{label\}\}. \\
    & "\{\{text\}\}" is not labeled as a \{\{label\}\} sentence. \\
    & This sentence, "\{\{text\}\}", is not classified as \{\{label\}\}. \\
    & The text "\{\{text\}\}" does not belong to the \{\{label\}\} category. \\
    & "\{\{text\}\}" is not a sentence of the \{\{label\}\} type. \\
\bottomrule
\end{tabular}
}

\subsection{Paraphrase Detection Templates}
\subsubsection{PAWS-X Templates}

\resizebox{\columnwidth}{!}{
\begin{tabular}{ll}\toprule
\textbf{Task} &\textbf{Statement Template} \\\midrule
\multirow{14}{*}{PAWS-X} & "\{\{text1\}\}" can be stated as "\{\{text2\}\}". \\
    & "\{\{text1\}\}" can not be stated as "\{\{text2\}\}". \\
    & "\{\{text1\}\}" can't be stated as "\{\{text2\}\}". \\
    & "\{\{text1\}\}" duplicates "\{\{text2\}\}". \\
    & "\{\{text1\}\}" does not duplicate "\{\{text2\}\}". \\
    & "\{\{text1\}\}" doesn't duplicate "\{\{text2\}\}". \\
    & "\{\{text1\}\}" is a duplicate of "\{\{text2\}\}". \\
    & "\{\{text1\}\}" is not a duplicate of "\{\{text2\}\}". \\
    & "\{\{text1\}\}" is the same as "\{\{text2\}\}". \\
    & "\{\{text1\}\}" is not the same as "\{\{text2\}\}". \\
    & "\{\{text1\}\}" is unrelated to "\{\{text2\}\}". \\
    & "\{\{text1\}\}" is a paraphrase of "\{\{text2\}\}". \\
    & "\{\{text1\}\}" is not a paraphrase of "\{\{text2\}\}". \\
    & "\{\{text1\}\}" isn't a paraphrase of "\{\{text2\}\}". \\
\bottomrule
\end{tabular}
}

\subsection{Sentence Completion Templates}
\subsubsection{XCOPA Templates}

\resizebox{\columnwidth}{!}{
\begin{tabular}{ll}\toprule
\textbf{Task} &\textbf{Statement Template} \\\midrule
\multirow{6}{*}{XCOPA} & The cause of \{\{premise\}\} is that \{\{choice1/choice2\}\}. \\
    & \{\{premise\}\} due to \{\{choice1/choice2\}\}. \\
    & The effect of \{\{premise\}\} is that \{\{choice1/choice2\}\}. \\
    & \{\{premise\}\} therefore \{\{choice1/choice2\}\}. \\
    & \{\{premise\}\}, so \{\{choice1/choice2\}\}. \\
\bottomrule
\end{tabular}
}

\subsubsection{XStoryCloze Templates}
\resizebox{\columnwidth}{!}{
\begin{tabular}{ll}\toprule
\textbf{Task} &\textbf{Statement Template} \\\midrule
\multirow{10}{*}{XStoryCloze} & \{\{text1\}\} entails \{\{text2\}\}. \\
    & \{\{text1\}\}? yes, \{\{text2\}\}. \\
    & Premise: \{\{text1\}\}, Hypothesis: \{\{text2\}\}, label: Entailment. \\
    & \{\{text1\}\} is neutral with regards to \{\{text2\}\}. \\
    & \{\{text1\}\}? maybe, \{\{text2\}\}. \\
    &  Premise: \{\{text1\}\}, Hypothesis: \{\{text2\}\}, label: Neutral. \\
    & \{\{text1\}\} contradicts \{\{text2\}\}. \\
    &  \{\{text1\}\}? no, \{\{text2\}\}. \\
    & Premise: \{\{text1\}\}, Hypothesis: \{\{text2\}\}, label: Contradiction. \\
\bottomrule
\end{tabular}
}

\subsection{Natural Language Inference Templates}
\subsubsection{XNLI Templates}

\resizebox{\columnwidth}{!}{
\begin{tabular}{ll}\toprule
\textbf{Task} &\textbf{Statement Template} \\\midrule
\multirow{5}{*}{XNLI} & In \{\{sentence\}\}, \_ is: \{\{option1/option2\}\}. \\
    & Q:\{\{sentence\}\}, A: \{\{option1/option2\}\}. \\
    & The missing word in \{\{sentence\}\} is \{\{option1/option2\}\}. \\
    & \_ in: \{\{sentence\}\} is \{\{option1/option2\}\}. \\
    & \{\{sentence\}\}, \_ is: \{\{option1/option2\}\}. \\
\bottomrule
\end{tabular}
}

\subsection{Coreference Resolution Templates}
\subsubsection{XWinograd Templates}

\resizebox{\columnwidth}{!}{
\begin{tabular}{ll}\toprule
\textbf{Task} &\textbf{Statement Template} \\\midrule
\multirow{3}{*}{XWinograd} & \{\{input\_sentence\_1\}\} \{\{input\_sentence\_2\}\} \{\{input\_sentence\_3\}\} \{\{input\_sentence\_4\}\} The right way to close this story is: \{\{sentence\_quiz1/sentence\_quiz2\}\}. \\
    & \{\{input\_sentence\_1\}\} \{\{input\_sentence\_2\}\} \{\{input\_sentence\_3\}\} \{\{input\_sentence\_4\}\} The proper ending to this story is: \{\{sentence\_quiz1/sentence\_quiz2\}\}. \\
    & \{\{input\_sentence\_1\}\} \{\{input\_sentence\_2\}\} \{\{input\_sentence\_3\}\} \{\{input\_sentence\_4\}\} The correct ending to this story is: \{\{sentence\_quiz1/sentence\_quiz2\}\}. \\

\bottomrule
\end{tabular}
}

\section{Training Datasets}
\label{sec:training-tasks}
The following datasets where used to create the statement dataset of \mst~ and the instruction dataset for the decoder models: Belebele (reading comprehension) \citep{bandarkar2024belebelebenchmarkparallelreading}, Exams (Question Answering) \citep{hardalov-etal-2020-exams}, xQuAD (Question Answering) \citep{artetxe-etal-2020-cross} for multiple-choice question answering; WikiLingua \citep{ladhak-etal-2020-wikilingua} for summarization; FLORES-101 \citep{goyal-etal-2022-flores} for machine translation; Multilingual Sentiments (IndoNLU, Multilingual
Amazon Reviews, GoEmotions, Offenseval Dravidian, SemEval-2018 Task 1:
Affect in Tweets, Emotion, IMDB, Amazon Polarity, Yelp Reviews, Yelp Polarity) \citep{wilie2020indonlu, marc_reviews, demszky2020goemotions, chakravarthi-etal-2021-lre, dravidianoffensive-eacl, hande-etal-2020-kancmd, chakravarthi-etal-2020-corpus, chakravarthi-etal-2020-sentiment,  SemEval2018Task1, saravia-etal-2018-carer, maas-EtAl:2011:ACL-HLT2011, 10.1145/2507157.2507163, NIPS2015_250cf8b5, zhangCharacterlevelConvolutionalNetworks2015} for sentiment analysis; XL-WiC \citep{raganato-etal-2020-xl} for word sense disambiguation; MASSIVE \citep{fitzgerald-etal-2023-massive} for intent classification; Multilingual Fig-QA \citep{kabra-etal-2023-multi}, X-CSQA and X-CODAH \citep{lin-etal-2021-common} for commonsense reasoning; SIB-200 \citep{adelani-etal-2024-sib} for topic classification; and PAWS-X \citep{yang-etal-2019-paws} for paraphrase detection.

\newpage
\onecolumn
\section{Languages}
\label{sec:appendixi}

\begin{table*}[h]
    \small
    \centering
    \begin{tabular}{cccccc}
    \toprule
    \textbf{ISO} & \textbf{Language} & \textbf{Family} & \textbf{Subgrouping} & \textbf{Script} & \textbf{Resource} \\
    \midrule
    af & Afrikaans & Indo-European & Germanic & Latin & High \\
    ar & Arabic & Afro-Asiatic & Semitic & Arabic & High \\
    de & German & Indo-European & Germanic & Latin & High \\
    en & English & Indo-European & Germanic & Latin & High \\ 
    es & Spanish & Indo-European & Italic & Latin & High \\
    fr & French & Indo-European & Italic & Latin & High \\
    ga & Irish & Indo-European & Celtic & Latin & Low \\
    gu & Gujarati & Indo-European & Indo-Aryan & Gujarati & Low \\
    ha & Hausa & Afro-Asiatic & Chadic & Latin & Low \\
    hi & Hindi & Indo-European & Indo-Aryan & Devanagari & High\\
    id & Indonesian & Austronesian & Malayo-Polynesian & Latin & High \\
    ig & Igbo & Atlantic-Congo & Benue-Congo & Latin & Low \\
    is & Icelandic & Indo-European & Germanic & Latin & High \\
    it & Italian & Indo-European & Italic & Latin & High \\
    kk & Kazakh & Turkic & Common Turkic & Cyrillic & High \\
    ky & Kyrgyz & Turkic & Common Turkic & Cyrillic & Low \\
    lo & Lao & Tai-Kadai & Kam-Tai & Lao & Low \\
    mt & Maltese & Afro-Asiatic & Semitic & Latin & High \\
    ny & Nyanja & Atlantic-Congo & Benue-Congo & Latin & Low \\
    pt & Portuguese & Indo-European & Italic & Latin & High \\
    ru & Russian & Indo-European & Balto-Slavic & Cyrillic & High \\
    si & Sinhala & Indo-European & Indo-Aryan & Sinhala & Low \\
    tr & Turkish & Turkic & Common Turkic & Latin & High \\
    vi & Vietnamese & Austroasiatic & Vietic & Latin & High \\
    zh & Chinese & Sino-Tibetan & Sinitic & Han & High \\
    \bottomrule
    \end{tabular}
    \caption{Languages used in this study in alphabetical order of ISO 639-1 Code. Information on language family, subgrouping, script, and resource level is drawn from \citep{nllbteam2022languageleftbehindscaling}.}
    \label{tab:languages}
\end{table*}

\newpage

\section{Finetuning Setup}
\label{sec:appendixb}

We include the finetuning setup of the Statement Tuned Encoder models in Table~\ref{tab:finetuning-setup-encoders}. 
\begin{table*}[h]
    \centering
    \resizebox{\linewidth}{!}{
    \begin{tabular}{lrrrrrr}\toprule
\textbf{Model} &\textbf{\#Epochs} &\textbf{Batch Size} &\textbf{Learning Rate} &\textbf{Weight Decay} &\textbf{Warmup Ratio} \\\midrule
google-bert/bert-base-multilingual-cased (mBERT) &20 &\multirow{4}{*}{16} &1.00e-6 &\multirow{4}{*}{0.1} &\multirow{4}{*}{0.1} \\
microsoft/mdeberta-v3-base (mDeBERTa) &\multirow{3}{*}{15} & &2.00e-6 & & \\
FacebookAI/xlm-roberta-base (XLM-R base) & & &1.00e-6 & & \\
FacebookAI/xlm-roberta-large (XLM-R base) & & &2.00e-6 & & \\
\bottomrule
\end{tabular}}
    \caption{Finetuning Setup and Hyperparameters for each encoder model.}
    \label{tab:finetuning-setup-encoders}
\end{table*}

Additionally, the decoder models were Instruction finetuned. All models above 2B parameters are finetuned using QLoRA \cite{qlora}, while all models under 2B parameters are finetuned using full finetuning. We include the specific hyperparameters in Table~\ref{tab:finetuning-setup-decoders}. We used a custom instruction dataset of 150K examples constructed from the same task mixture as \st. The instruction templates are outlined in Appendix~\ref{sec:instruction-templates}.

\begin{table*}[h]
    \centering
    \resizebox{0.8\linewidth}{!}{
    \begin{tabular}{lrrrrrrr}\toprule
\textbf{Model} &\textbf{\#Epochs} &\textbf{Mode} &\textbf{Batch Size} &\textbf{Learning Rate} &\textbf{Weight Decay} &\textbf{Warmup Ratio} \\\midrule
Llama3.1 8B &\multirow{10}{*}{1} &\multirow{7}{*}{QLoRA} &\multirow{10}{*}{4} &0.00001 &\multirow{10}{*}{0} &steps=10 \\
Qwen2 72B & & & &0.0002 & &steps=10 \\
Llama3.1 70B & & & &0.00001 & &steps=10 \\
Gemma 2 9B & & & &0.0002 & &0.1 \\
Gemma 2 27B & & & &0.0002 & &0.1 \\
Aya 23 8b & & & &0.00001 & &steps=10 \\
Aya 23 35b & & & &0.00001 & &steps=10 \\
Gemma 2 2b & &\multirow{3}{*}{FFT} & &0.0002 & &steps=10 \\
Qwen2 1.5B & & & &0.0002 & &steps=10 \\
Qwen2 0.5B & & & &0.0002 & &steps=10 \\
\bottomrule
\end{tabular}}
    \caption{Finetuning Setup and Hyperparameters for each decoder model.}
    \label{tab:finetuning-setup-decoders}
\end{table*}

\twocolumn
\section{Language Level Performance}
\label{sec:lang-level}
\begin{figure*}[h]
 \centering
 \begin{subfigure}{0.47\textwidth}
     \centering
     \includegraphics[width=\textwidth]{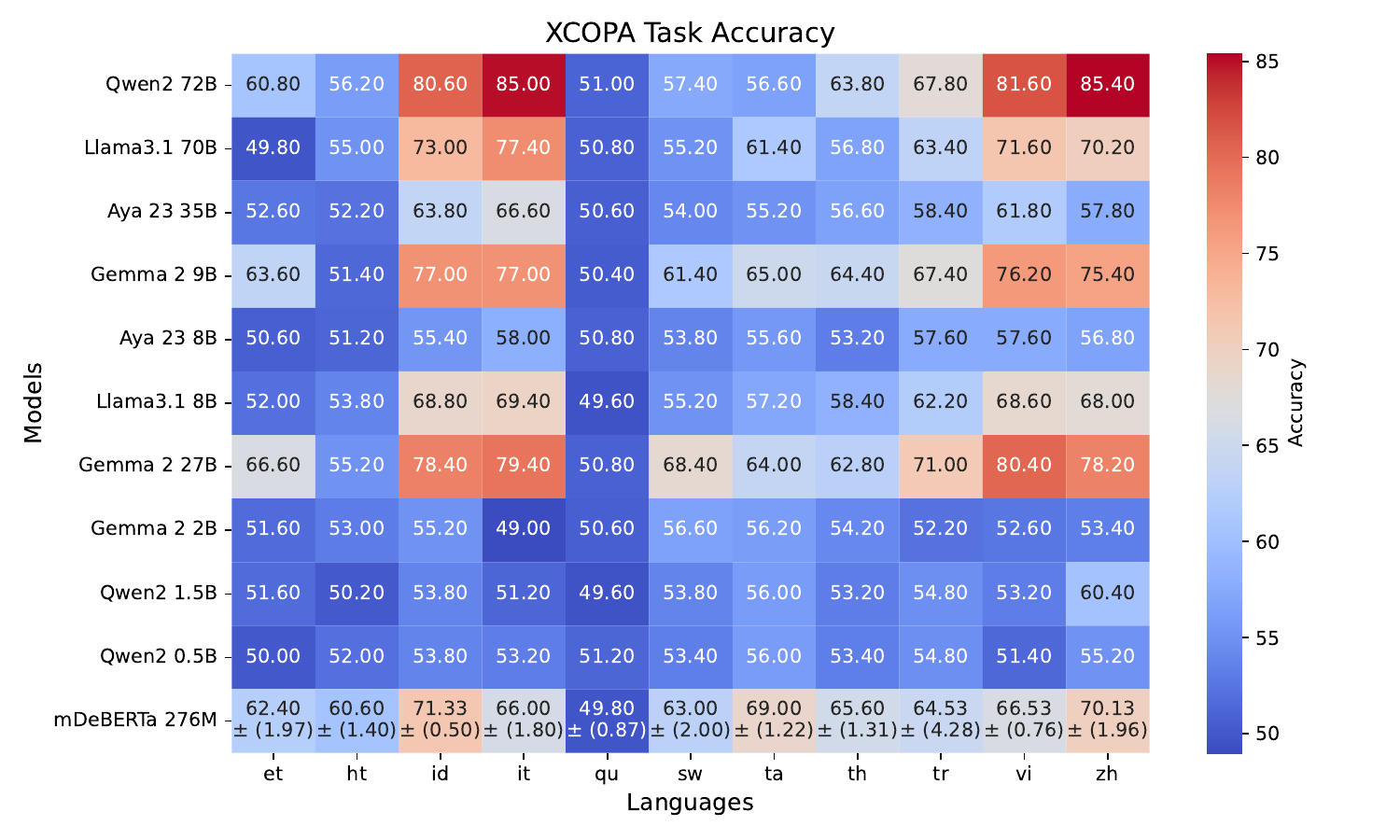}
     \label{fig:a}
 \end{subfigure}
 \hfill
 \begin{subfigure}{0.47\textwidth}
     \centering
     \includegraphics[width=\textwidth]{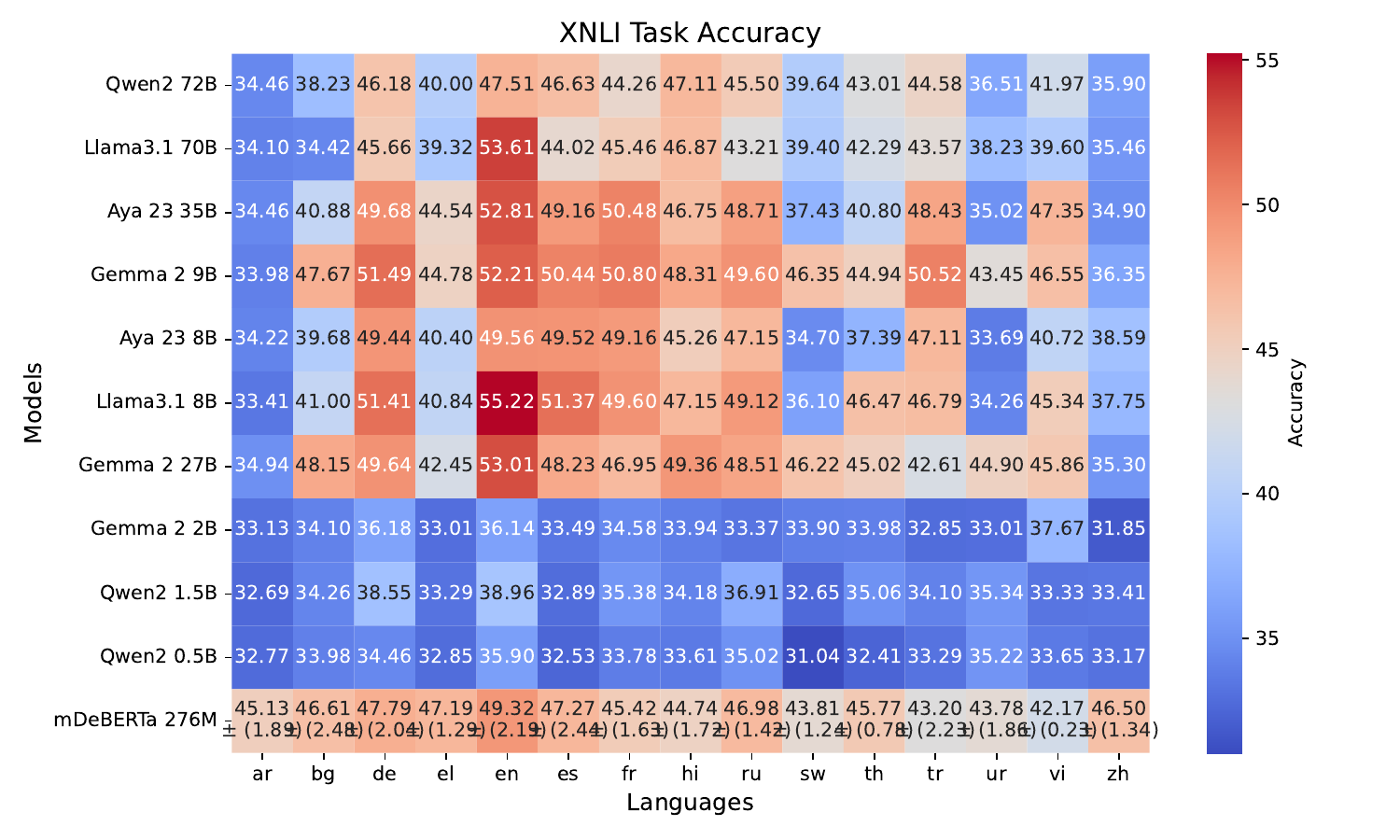}
     \label{fig:b}
 \end{subfigure}
 
 \bigskip
 
 \begin{subfigure}{0.47\textwidth}
     \centering
     \includegraphics[width=\textwidth]{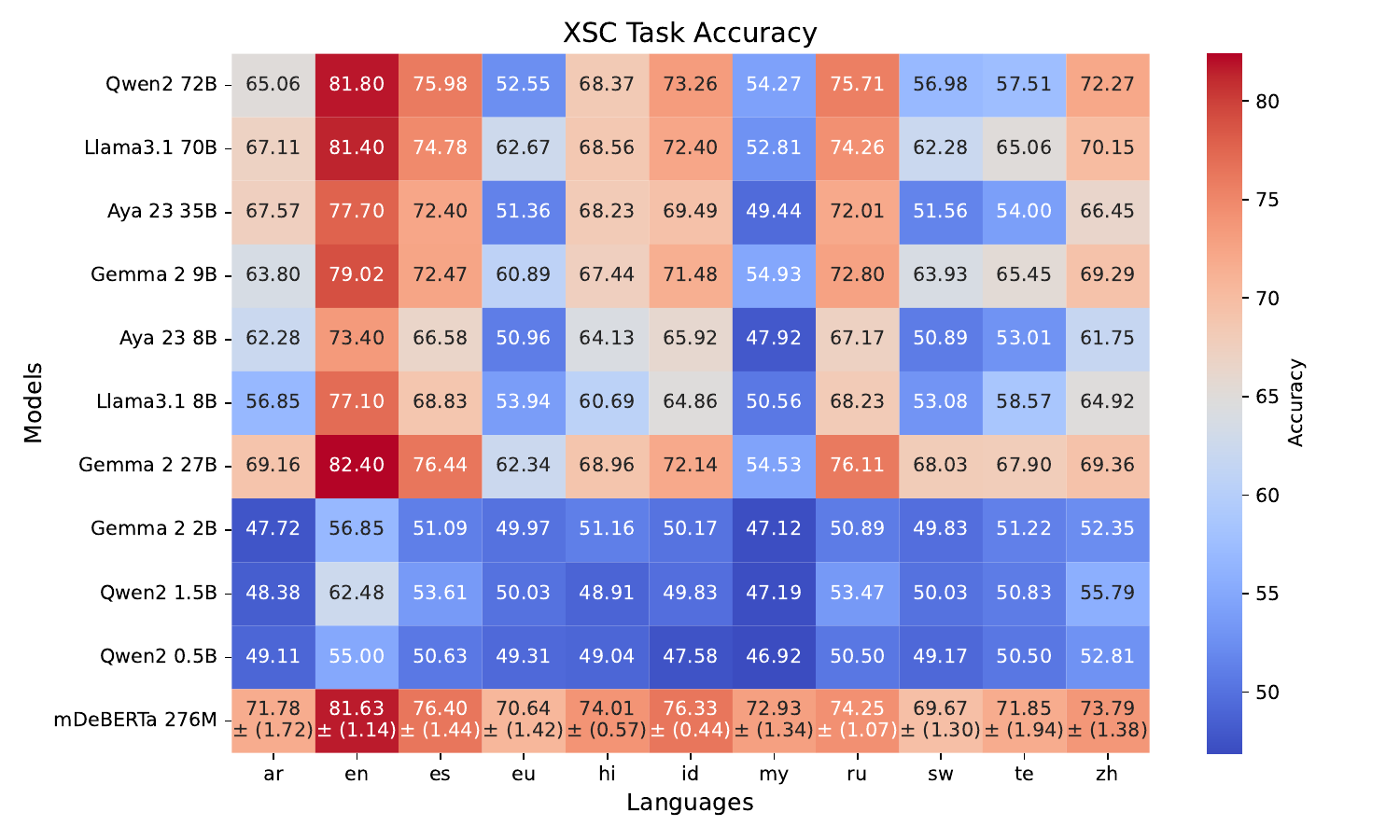}
     \label{fig:c}
 \end{subfigure}
 \hfill
 \begin{subfigure}{0.47\textwidth}
     \centering
     \includegraphics[width=\textwidth]{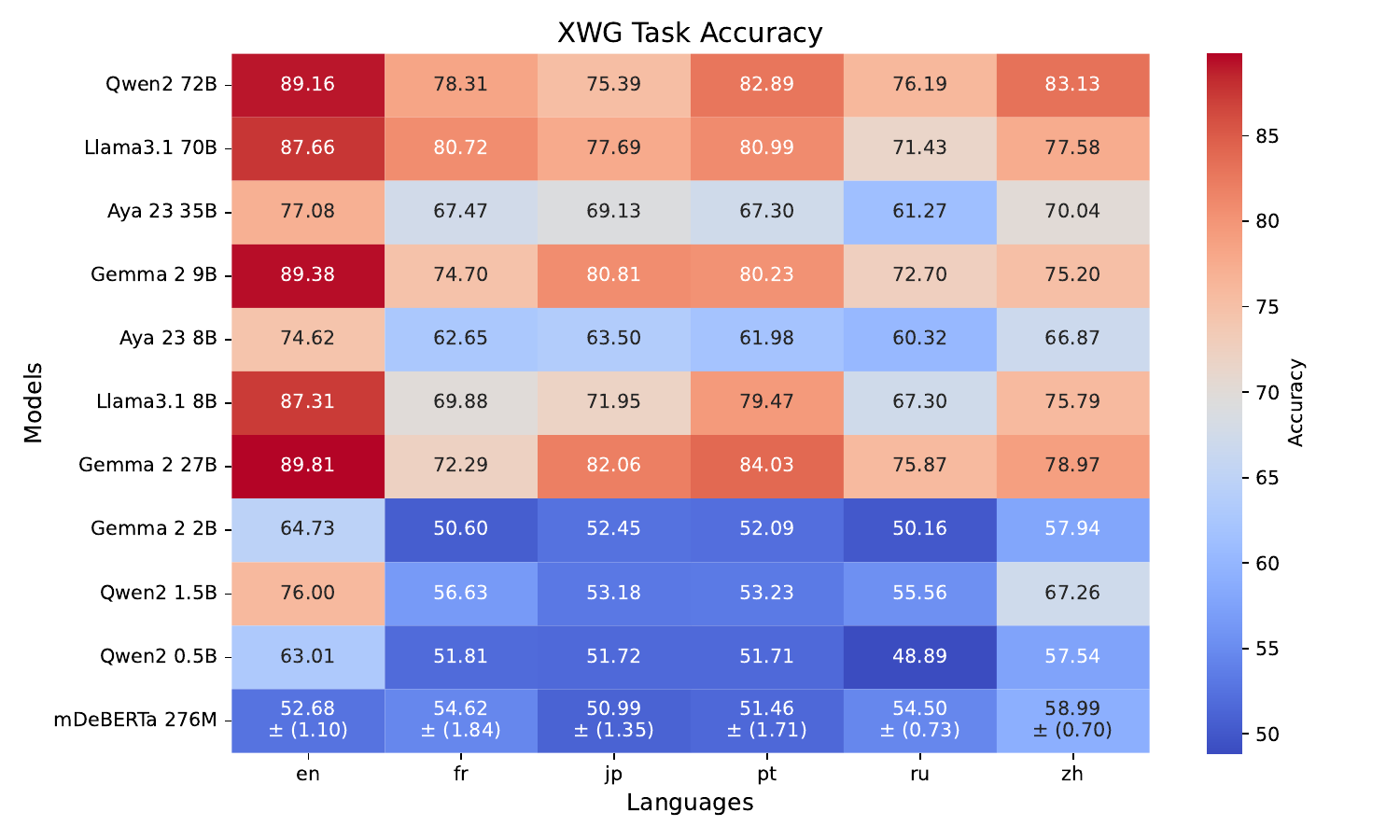}
     \label{fig:d}
 \end{subfigure}
\caption{Individual language subset performance on all 4 evaluation tasks and decoder models finetuned on the same data mixture as \st and mDeBERTa trained on 11-langs. We include the standard deviation in performance over 3 training runs for mDeBERTa.}
 \label{fig:individual-lang}
\end{figure*}

\begin{figure*}[h]
 \centering
 \begin{subfigure}{0.47\textwidth}
     \centering
     \includegraphics[width=\textwidth]{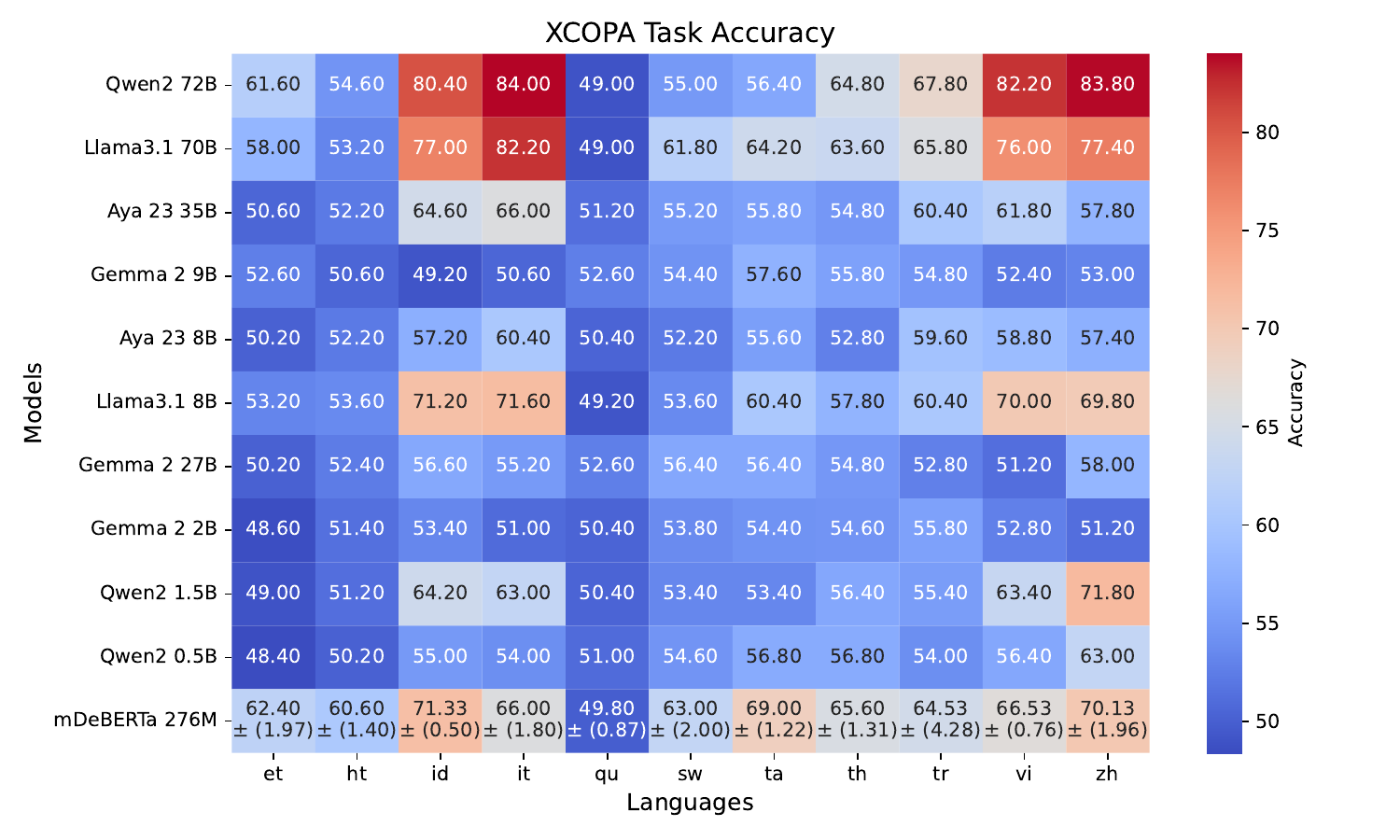}
     \label{fig:a_old}
 \end{subfigure}
 \hfill
 \begin{subfigure}{0.47\textwidth}
     \centering
     \includegraphics[width=\textwidth]{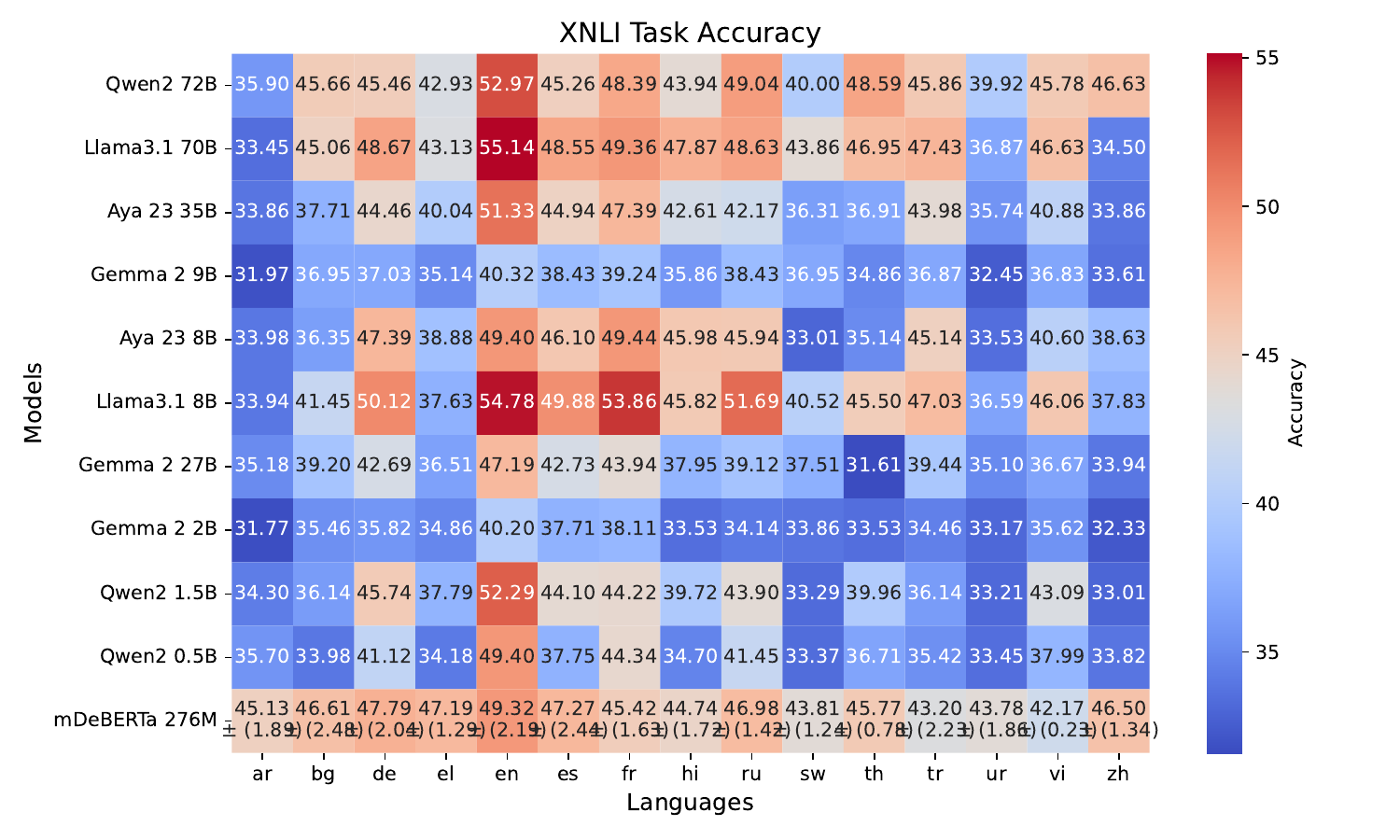}
     \label{fig:b_old}
 \end{subfigure}
 
 \bigskip
 
 \begin{subfigure}{0.47\textwidth}
     \centering
     \includegraphics[width=\textwidth]{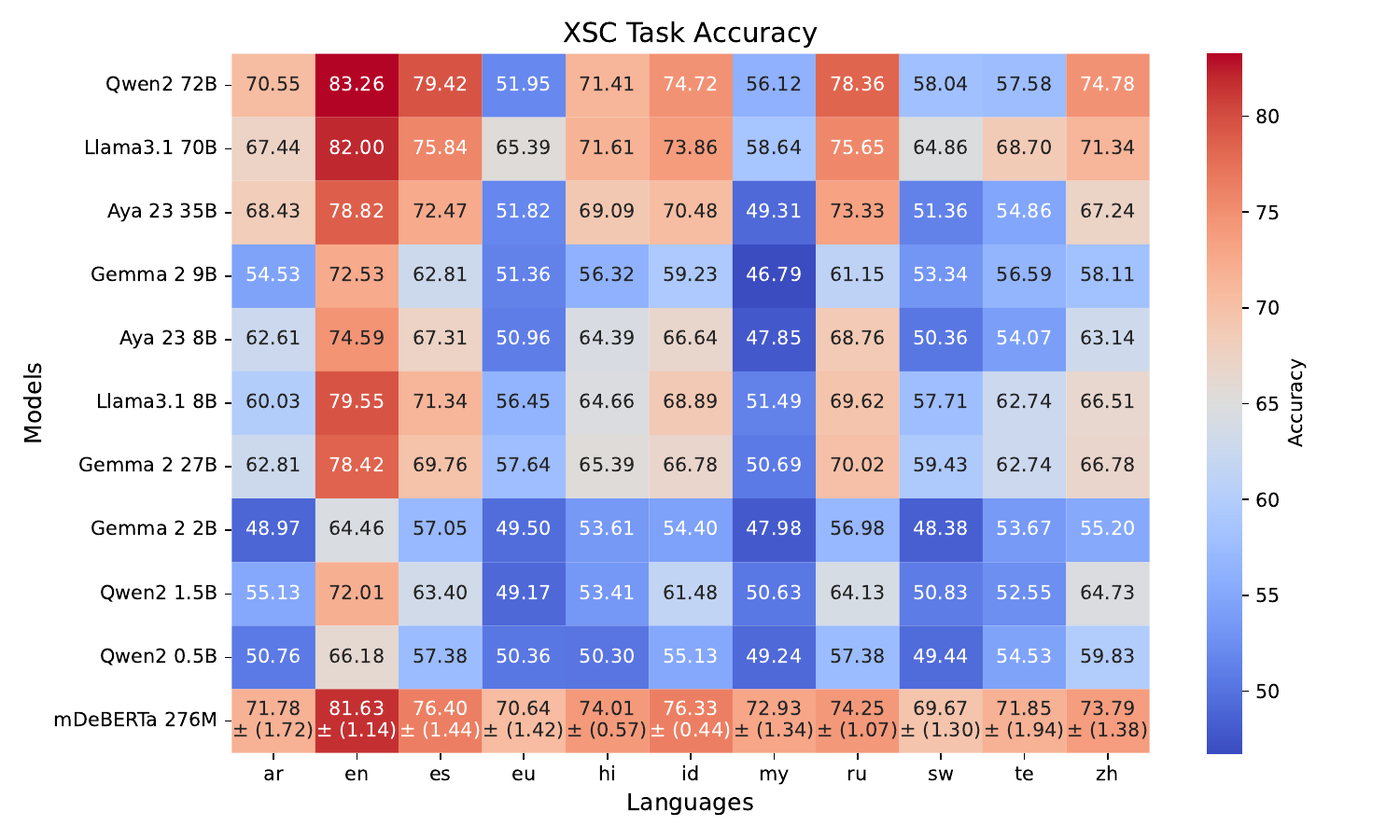}
     \label{fig:c_old}
 \end{subfigure}
 \hfill
 \begin{subfigure}{0.47\textwidth}
     \centering
     \includegraphics[width=\textwidth]{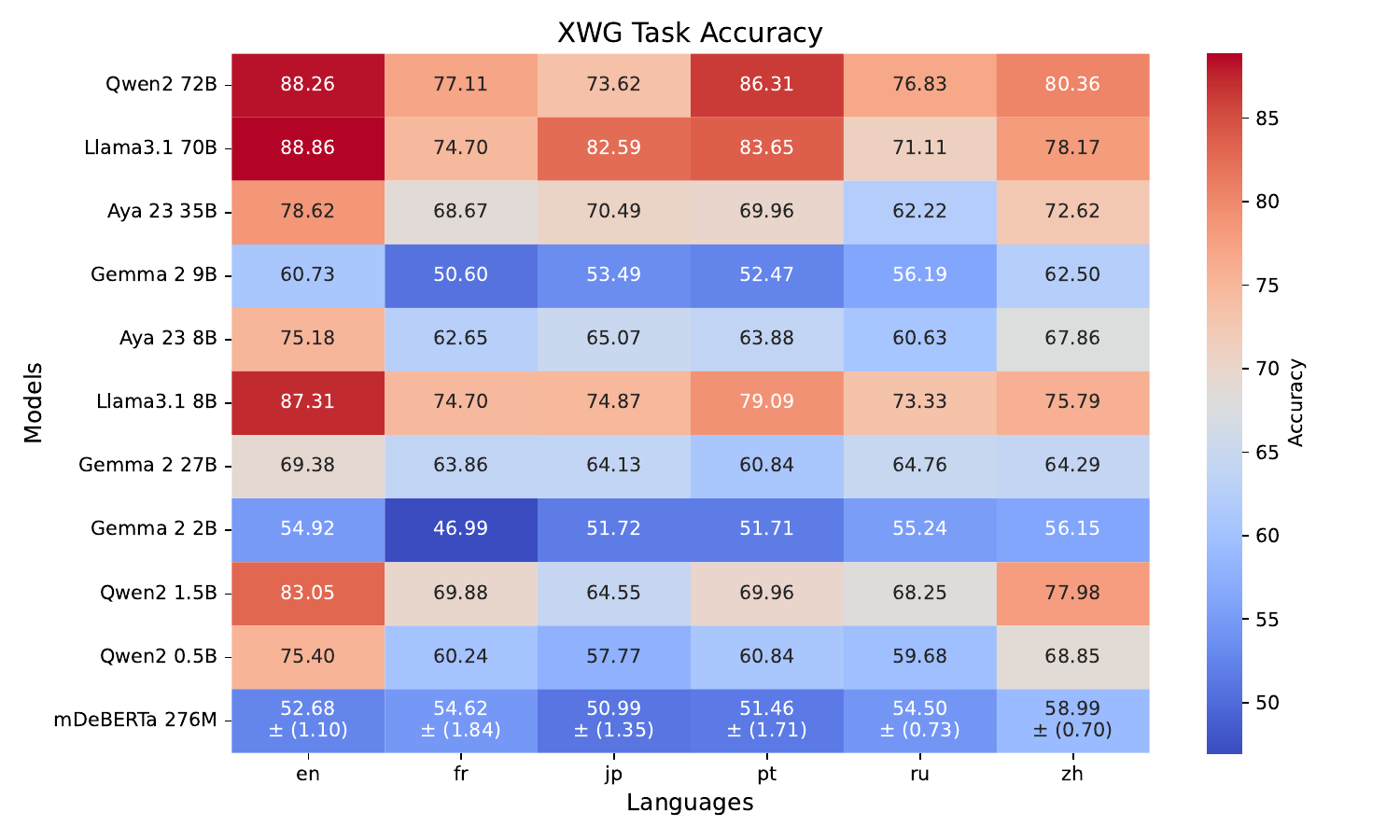}
     \label{fig:d_old}
 \end{subfigure}

 \caption{Individual language subset performance on all 4 evaluation tasks and decoder models (instruction-tuned on custom data by the teams who released the models) and mDeBERTa trained on 11-langs. We include the standard deviation in performance over 3 training runs for mDeBERTa.}
 \label{fig:individual-lang-old}
\end{figure*}

In Figure~\ref{fig:individual-lang} (fine-tuned on the same data) and Figure~\ref{fig:individual-lang-old} (fine-tuned on custom data mixtures by the teams who developed the models), we report the individual language performance on all 4 evaluation tasks of all generative models and mDeBERTa finetuned using \st. There are several interesting observations.

In both cases of instruction finetuning setup, we observe largely the same trends. First, on XCOPA, XNLI, and XStoryCloze we notice that mDeBERTa tends to perform more equitably than the LLMs, meaning that there is less variation \textbf{across} languages in a single task/dataset. For example, in XCOPA, Qwen2 72B and Llama3.1 70B perform strongly on Indonesian, Italian, Vietnamese, and Chinese but have lackluster performance on most of the other languages. While mDeBERTa seems to have less deviation between the best performing languages and the others. We see this in XNLI and XStoryCloze as well (for example Arabic, Swahili, and Urdu in XNLI, and Swahilli, Telugu, and Basque in StoryCloze). This adds more support for the use of our method with lower resource/tail-end languages.

Second, we notice that our method can generalize to languages/language families that are unseen during \st~ if they are seen during pretraining. For example, Turkish (tr) and all Turkic languages for that matter are completely unseen during \st, but are seen during pretraining, the model was still able to generalize on Turkish on XNLI performing on par with Aya 23 35b. Moreover, Burmese (my) and all its closely related languages  are completely unseen during \st~ while being seen during pretraining, but the performance on XStoryCloze was exceptionally strong far outperforming even the strongest generative model (\textbf{72.93} of mDeBERTa vs. \textbf{52.81} of Llama3.1 70B fine-tuned on the same data mixture). On the other hand, a language where our model fails to generalize Quechua (qu) on XCOPA was completely unseen during \st~and pretraining. This is encouraging as it further supports our hypothesis that multilingual pretraining is what powers Multilingual \st. This should encourage the development of more powerful encoder-only models with support for more languages.

\section{Languages Used during Statement Tuning}
\label{language-outline}
As a subset of the potential 25 languages from the training set we choose the following 11 languages as an intermediate subset:

\begin{itemize}
    \itemsep0em
    \item Chinese
    \item English
    \item French
    \item Vietnamese
    \item Swahili
    \item Russian
    \item Arabic
    \item Hindi
    \item German
    \item Indonesian
    \item Italian
\end{itemize}

We make the choice of these specific languages as they span a variety of language families, scripts, and resource availability and hence could potentially help with cross-lingual generalization. 

\onecolumn
\section{Performance of Instruction-Tuned Model Variants}
\label{sec:Instruction-Tuned}
\begin{table*}[ht!]
\centering
\resizebox{0.8\linewidth}{!}{
\begin{tabular}{lcccccc}
\toprule
\textbf{Model} & \textbf{Parameters} & \textbf{XCOPA} & \textbf{XNLI} & \textbf{XStoryCloze} & \textbf{XWinoGrad} \\
\midrule
Qwen2 & 72B & 67.24 & \textcolor{lightgray}{\textbf{45.09}} & \textcolor{lightgray}{\textbf{68.74}} & 80.42 \\
Llama3.1 & 70B & 66.20 & \textcolor{lightgray}{\textbf{45.07}} & \textcolor{lightgray}{\textbf{70.48}} & 79.85  \\
Gemma 2 & 9B & \textcolor{lightgray}{\textbf{53.00}} & \textcolor{lightgray}{\textbf{36.33 }}& \textcolor{lightgray}{\textbf{57.52}} & 56.00  \\
Llama3.1 & 8B & \textcolor{lightgray}{\textbf{60.98}} & \textcolor{lightgray}{\textbf{44.84}} & \textcolor{lightgray}{\textbf{64.45}} & 77.52  \\
Aya 23 & 8B & \textcolor{lightgray}{\textbf{55.16}} & \textcolor{lightgray}{\textbf{41.30}} & \textcolor{lightgray}{\textbf{60.97}} & 65.88  \\
Aya 23 & 35B & \textcolor{lightgray}{\textbf{57.31}} & \textcolor{lightgray}{\textbf{40.81}} & \textcolor{lightgray}{\textbf{64.29}} & 70.43  \\
Gemma 2 & 27B & \textcolor{lightgray}{\textbf{54.24}} & \textcolor{lightgray}{\textbf{38.59}} & \textcolor{lightgray}{\textbf{64.59}} & 64.54 \\
Gemma 2 & 2B & \textcolor{lightgray}{\textbf{52.49}} & \textcolor{lightgray}{\textbf{34.97 }}& \textcolor{lightgray}{\textbf{53.65}} & 52.79  \\
Qwen2 & 1.5B & \textcolor{lightgray}{\textbf{57.42}} &\textcolor{lightgray}{\textbf{ 39.79}} & \textcolor{lightgray}{\textbf{57.95}} & 72.28 \\
Qwen2 & 500M & \textcolor{lightgray}{\textbf{54.56}} &\textcolor{lightgray}{\textbf{ 37.56}} & \textcolor{lightgray}{\textbf{54.59}} & 63.80 \\

\midrule
mBERT(base) & 110M & 52.47 & 34.51 & 48.30 & 50.68  \\
XLMR-base & 250M & 56.69 & 35.33 & 60.71 & 51.34  \\
XLMR-large & 560M & 64.36 & 45.76 & 78.78 & 54.26  \\
\textbf{mDeBERTa (Best)} & 276M & 65.52$_{(1.64)}$ & 47.84$_{(1.65)}$ & 73.53$_{(1.25)}$ & 54.75$_{(1.24)}$  \\
\bottomrule
\end{tabular}
}
\caption{\textbf{Accuracy of the existing instruction-tuned varieities of multilingual decoder and Statement-Tuned encoder models} on XCOPA, XNLI, XStoryCloze, and XWinoGrad tasks. Results in grey highlight performances that are below the best-performing encoder model, mDeBERTa (276M). Additionally, we report the average standard deviation across languages over 3 training runs only for mDeBERTa to quantify the random deviation due to \st~training.}
\label{fig:decoder-baselines-old}
\end{table*}

\clearpage
\twocolumn
\section{Instruction Templates}
\label{sec:instruction-templates}
\subsection{Multiple-Choice QA Templates}
\subsubsection{Belebele Templates}

\renewcommand{\arraystretch}{}
\resizebox{\columnwidth}{!}{
\begin{tabular}{ll}\toprule
\textbf{Task} &\textbf{Instruction Template} \\\midrule
\multirow{10}{*}{Belebele} & \{\{flores\_passage\}\}\textbackslash n\textbackslash nQuestion: \{\{question\}\}\textbackslash n\{\{options\_\}\} \\
& \{\{flores\_passage\}\}\textbackslash n\textbackslash n\{\{question\}\}\textbackslash n\{\{options\_\}\} \\
& \{\{flores\_passage\}\}\textbackslash n\textbackslash nAnswer the following question: \{\{question\}\}\textbackslash n\{\{options\_\}\} \\
& \{\{flores\_passage\}\}\textbackslash n\textbackslash nBased on the preceding passage, answer the following question \{\{question\}\}\textbackslash n\{\{options\_\}\} \\
& \{\{flores\_passage\}\}\textbackslash n\textbackslash nGive an answer to the following question using evidence from the above passage: \{\{question\}\}\textbackslash n\{\{options\_\}\} \\
& Context: \{\{flores\_passage\}\}\textbackslash nQuestion \{\{question\}\}\textbackslash nAnswer:\textbackslash n\{\{options\_\}\} \\
& Read the following passage and answer the question.\textbackslash n\textbackslash n\{\{flores\_passage\}\}\textbackslash n\textbackslash n\{\{question\}\}\textbackslash n\{\{options\_\}\} \\
& Answer the question about the text:\textbackslash n\textbackslash n\{\{flores\_passage\}\}\textbackslash n\textbackslash n\{\{question\}\}\textbackslash n\{\{options\_\}\} \\
& \{\{flores\_passage\}\}\textbackslash n\textbackslash nWhat is the correct answer to the following question based on the text?\textbackslash n\{\{question\}\}\textbackslash n\{\{options\_\}\} \\
& Refer to the passage below:\textbackslash n\textbackslash n\{\{flores\_passage\}\}\textbackslash n\textbackslash nWhat is the answer to this question?\textbackslash n\{\{question\}\}\textbackslash n\{\{options\_\}\} \\
\bottomrule
\end{tabular}
}

\subsubsection{Exams Templates}

\resizebox{\columnwidth}{!}{
\begin{tabular}{ll}\toprule
\textbf{Task} &\textbf{Instruction Template} \\\midrule
    \multirow{10}{*}{Exams} & Question: \{\{stem\}\}\textbackslash nChoices:\textbackslash nA. \{\{choice1\}\}\textbackslash nB. \{\{choice2\}\}\textbackslash nC. \{\{choice3\}\}\textbackslash nD. \{\{choice4\}\} \\
    & \{\{stem\}\}\textbackslash n\textbackslash nOptions:\textbackslash nA. \{\{choice1\}\}\textbackslash nB. \{\{choice2\}\}\textbackslash nC. \{\{choice3\}\}\textbackslash nD. \{\{choice4\}\} \\
    & Read the question:\textbackslash n\textbackslash n\{\{stem\}\}\textbackslash n\textbackslash nChoices:\textbackslash nA. \{\{choice1\}\}\textbackslash nB. \{\{choice2\}\}\textbackslash nC. \{\{choice3\}\}\textbackslash nD. \{\{choice4\}\} \\
    & What is the correct answer to this question?\textbackslash n\textbackslash n\{\{stem\}\}\textbackslash n\textbackslash nOptions:\textbackslash nA. \{\{choice1\}\}\textbackslash nB. \{\{choice2\}\}\textbackslash nC. \{\{choice3\}\}\textbackslash nD. \{\{choice4\}\} \\
    & Based on the question below, choose the most accurate answer:\textbackslash n\textbackslash n\{\{stem\}\}\textbackslash n\textbackslash nChoices:\textbackslash nA. \{\{choice1\}\}\textbackslash nB. \{\{choice2\}\}\textbackslash nC. \{\{choice3\}\}\textbackslash nD. \{\{choice4\}\} \\
    & Select the right option for the following question:\textbackslash n\textbackslash n\{\{stem\}\}\textbackslash n\textbackslash nA. \{\{choice1\}\}\textbackslash nB. \{\{choice2\}\}\textbackslash nC. \{\{choice3\}\}\textbackslash nD. \{\{choice4\}\} \\
    & Question:\textbackslash n\{\{stem\}\}\textbackslash n\textbackslash nOptions:\textbackslash n1. \{\{choice1\}\}\textbackslash n2. \{\{choice2\}\}\textbackslash n3. \{\{choice3\}\}\textbackslash n4. \{\{choice4\}\} \\
    & Answer the following question by selecting the best choice:\textbackslash n\textbackslash n\{\{stem\}\}\textbackslash n\textbackslash nA. \{\{choice1\}\}\textbackslash nB. \{\{choice2\}\}\textbackslash nC. \{\{choice3\}\}\textbackslash nD. \{\{choice4\}\} \\
    & \{\{stem\}\}\textbackslash n\textbackslash nAnswer options:\textbackslash n1. \{\{choice1\}\}\textbackslash n2. \{\{choice2\}\}\textbackslash n3. \{\{choice3\}\}\textbackslash n4. \{\{choice4\}\} \\
    & Below is a question and four possible answers. Choose one.\textbackslash n\textbackslash nQuestion: \{\{stem\}\}\textbackslash n\textbackslash nChoices:\textbackslash nA. \{\{choice1\}\}\textbackslash nB. \{\{choice2\}\}\textbackslash nC. \{\{choice3\}\}\textbackslash nD. \{\{choice4\}\} \\\bottomrule
\end{tabular}
}

\subsubsection{xQuAD Templates}

\resizebox{\columnwidth}{!}{
\begin{tabular}{ll}\toprule
\textbf{Task} &\textbf{Instruction Template} \\\midrule
    \multirow{10}{*}{xQuAD} & Please answer a question about the following article:\textbackslash n\textbackslash n\{\{context\}\}\textbackslash n\textbackslash n\{\{question\}\} \\
    & Read this and answer the question\textbackslash n\textbackslash n\{\{context\}\}\textbackslash n\textbackslash n\{\{question\}\} \\
    & \{\{context\}\}\textbackslash n\{\{question\}\} \\
    & Answer a question about this article:\textbackslash n\{\{context\}\}\textbackslash n\{\{question\}\} \\
    & Here is an article: \{\{context\}\}\textbackslash nWhat is the answer to this question: \{\{question\}\} \\
    & Article: \{\{context\}\}\textbackslash n\textbackslash nQuestion: \{\{question\}\} \\
    & Article: \{\{context\}\}\textbackslash n\textbackslash nNow answer this question: \{\{question\}\} \\
    & \{\{context\}\}\textbackslash n\textbackslash nQ: \{\{question\}\} \\
    & Read the following article and answer the question:\textbackslash n\textbackslash n\{\{context\}\}\textbackslash n\textbackslash nQuestion: \{\{question\}\} \\
    & Please read this passage and provide an answer to the following question:\textbackslash n\textbackslash n\{\{context\}\}\textbackslash n\textbackslash nQuestion: \{\{question\}\} \\\bottomrule
\end{tabular}
}

\subsection{Summarization Templates}
\subsubsection{WikiLingua Templates}

\resizebox{\columnwidth}{!}{
\begin{tabular}{ll}\toprule
\textbf{Task} &\textbf{Instruction Template} \\\midrule
    \multirow{10}{*}{WikiLingua} & Summarize:\textbackslash n\textbackslash n\{\{source\}\} \\
    & Summarize the following:\textbackslash n\{\{source\}\} \\
    & Summarize this passage:\textbackslash n\textbackslash n\{\{source\}\} \\
    & Provide a concise summary of this passage:\textbackslash n\{\{source\}\} \\
    & What is a summary of the following passage?\textbackslash n\{\{source\}\} \\
    & Describe the key points of the following passage:\textbackslash n\textbackslash n\{\{source\}\} \\
    & Passage: \{\{source\}\}\textbackslash n\textbackslash nWhat is a summary? \\
    & Passage: \{\{source\}\}\textbackslash nWhat is a summary of what this passage is about? \\
    & Provide a brief overview of the following passage:\textbackslash n\textbackslash n\{\{source\}\} \\
    & Condense the key information from the following passage:\textbackslash n\textbackslash n\{\{source\}\} \\\bottomrule
\end{tabular}

}

\subsection{Machine Translation Templates}
\subsubsection{FLORES-101 Templates}

\resizebox{\columnwidth}{!}{
\begin{tabular}{ll}\toprule
\textbf{Task} & \textbf{Instruction Template} \\\midrule
\multirow{10}{*}{FLORES-101} 
    & Translate the following sentence from \{\{ori\_lang\}\} to \{\{target\_lang\}\}:\textbackslash n\textbackslash n\{\{ori\_sen\}\} \\
    & Please translate this sentence from \{\{ori\_lang\}\} into \{\{target\_lang\}\}:\textbackslash n\textbackslash n\{\{ori\_sen\}\} \\
    & Translate the \{\{ori\_lang\}\} sentence to \{\{target\_lang\}\}:\textbackslash n\textbackslash n\{\{ori\_sen\}\} \\
    & Convert the following \{\{ori\_lang\}\} sentence into \{\{target\_lang\}\}:\textbackslash n\textbackslash n\{\{ori\_sen\}\} \\
    & Translate this sentence from \{\{ori\_lang\}\} into \{\{target\_lang\}\}:\textbackslash n\{\{ori\_sen\}\} \\
    & Please provide the translation of the following \{\{ori\_lang\}\} sentence into \{\{target\_lang\}\}:\textbackslash n\{\{ori\_sen\}\} \\
    & Translate this \{\{ori\_lang\}\} sentence to \{\{target\_lang\}\}:\textbackslash n\{\{ori\_sen\}\} \\
    & Convert this \{\{ori\_lang\}\} sentence to \{\{target\_lang\}\}:\textbackslash n\{\{ori\_sen\}\} \\
    & Given the following \{\{ori\_lang\}\} sentence, translate it into \{\{target\_lang\}\}:\textbackslash n\textbackslash n\{\{ori\_sen\}\} \\
    & How do you say the following sentence in \{\{target\_lang\}\}?\textbackslash n\{\{ori\_sen\}\} \\\bottomrule
\end{tabular}

}

\subsection{Sentiment Analysis Templates}
\subsubsection{multilingual-sentiments Templates}

\resizebox{\columnwidth}{!}{
\begin{tabular}{ll}\toprule
\textbf{Task} & \textbf{Instruction Template} \\\midrule
\multirow{10}{*}{multilingual-sentiments} 
    & \{\{text\}\}\textbackslash nWhat is the sentiment of this text? (positive, neutral, negative) \\
    & \{\{text\}\}\textbackslash nIs the sentiment of this text positive, neutral, or negative? \\
    & What sentiment does this text express (positive, neutral, negative)?\textbackslash n\textbackslash n\{\{text\}\} \\
    & Describe the sentiment of the following text (positive, neutral, negative):\textbackslash n\textbackslash n\{\{text\}\} \\
    & How would you classify the sentiment of this text?\textbackslash n\textbackslash n\{\{text\}\} (positive, neutral, negative) \\
    & What sentiment best describes this text? (positive, neutral, negative)\textbackslash n\textbackslash n\{\{text\}\} \\
    & Analyze the sentiment of the following text:\textbackslash n\textbackslash n\{\{text\}\} (positive, neutral, negative) \\
    & \{\{text\}\}\textbackslash nLabel the sentiment as positive, neutral, or negative: \\
    & Classify the sentiment expressed in this text as positive, neutral, or negative:\textbackslash n\textbackslash n\{\{text\}\} \\
    & Determine whether the sentiment of this text is positive, neutral, or negative:\textbackslash n\textbackslash n\{\{text\}\} \\\bottomrule
\end{tabular}

}

\subsection{Word Sense Disambiguation Templates}
\subsubsection{XL-WiC Templates}

\resizebox{\columnwidth}{!}{
\begin{tabular}{ll}\toprule
\textbf{Task} & \textbf{Instruction Template} \\\midrule
\multirow{10}{*}{XL-WiC} 
    & Context 1: \{\{context\_1\}\}\textbackslash nContext 2: \{\{context\_2\}\}\textbackslash nDoes the word '\{\{target\_word\}\}' have the same meaning in both contexts? (Yes or No) \\
    & Given the word '\{\{target\_word\}\}', compare the following two contexts:\textbackslash nContext 1: \{\{context\_1\}\}\textbackslash nContext 2: \{\{context\_2\}\}\textbackslash nIs the word used in the same sense in both contexts? (Yes or No) \\
    & Context 1: \{\{context\_1\}\}\textbackslash nContext 2: \{\{context\_2\}\}\textbackslash nDoes the target word '\{\{target\_word\}\}' have the same meaning in both contexts? (Yes or No) \\
    & Context 1: \{\{context\_1\}\}\textbackslash nContext 2: \{\{context\_2\}\}\textbackslash nIs the word '\{\{target\_word\}\}' used in the same sense in both contexts? (Yes or No) \\
    & Check if the word '\{\{target\_word\}\}' has the same meaning in the following contexts:\textbackslash nContext 1: \{\{context\_1\}\}\textbackslash nContext 2: \{\{context\_2\}\}\textbackslash nAnswer with 'Yes' or 'No' \\
    & Does the word '\{\{target\_word\}\}' have the same sense in both contexts?\textbackslash nContext 1: \{\{context\_1\}\}\textbackslash nContext 2: \{\{context\_2\}\}\textbackslash nAnswer 'Yes' or 'No' \\
    & Context 1: \{\{context\_1\}\}\textbackslash nContext 2: \{\{context\_2\}\}\textbackslash nDoes the word '\{\{target\_word\}\}' share the same meaning in both? (Yes/No) \\
    & Given the following contexts, is the meaning of '\{\{target\_word\}\}' the same in both?\textbackslash nContext 1: \{\{context\_1\}\}\textbackslash nContext 2: \{\{context\_2\}\}\textbackslash nAnswer 'Yes' or 'No' \\
    & Is the word '\{\{target\_word\}\}' used with the same meaning in both contexts?\textbackslash nContext 1: \{\{context\_1\}\}\textbackslash nContext 2: \{\{context\_2\}\}\textbackslash nAnswer 'Yes' or 'No' \\
    & Context 1: \{\{context\_1\}\}\textbackslash nContext 2: \{\{context\_2\}\}\textbackslash nDo the two contexts convey the same meaning for the word '\{\{target\_word\}\}'? (Yes/No) \\\bottomrule
\end{tabular}

}

\subsection{Intent Classification Templates}
\subsubsection{MASSIVE Templates}

\resizebox{\columnwidth}{!}{
\begin{tabular}{ll}\toprule
\textbf{Task} & \textbf{Instruction Template} \\\midrule
\multirow{10}{*}{MASSIVE} 
    & \{\{utt\}\}\textbackslash nWhat is the scenario of this utterance? (Choose from \{\{options\_\}\}) \\
    & Given the following utterance: \{\{utt\}\}\textbackslash nWhat scenario does it belong to? (Select from \{\{options\_\}\}) \\
    & \{\{utt\}\}\textbackslash nWhich scenario best describes this utterance? (Choose one from \{\{options\_\}\}) \\
    & What is the appropriate scenario for this utterance?\textbackslash n\{\{utt\}\}\textbackslash nSelect from the following scenarios: \{\{options\_\}\} \\
    & Context: \{\{utt\}\}\textbackslash nWhat scenario is being described? (Options: \{\{options\_\}\}) \\
    & Utterance: \{\{utt\}\}\textbackslash nDetermine the scenario of this utterance from the following options: \{\{options\_\}\} \\
    & \{\{utt\}\}\textbackslash nIn which scenario does this utterance fit? (Choose from \{\{options\_\}\}) \\
    & Based on the following utterance, classify the scenario: \{\{utt\}\}\textbackslash nPossible options: \{\{options\_\}\} \\
    & \{\{utt\}\}\textbackslash nIdentify the scenario that best fits this utterance (options: \{\{options\_\}\}) \\
    & What scenario does the following utterance belong to?\textbackslash n\{\{utt\}\}\textbackslash nAvailable options: \{\{options\_\}\} \\\bottomrule
\end{tabular}

}

\subsection{Commonsense Reasoning Templates}
\subsubsection{Multilingual Fig-QA Templates}

\resizebox{\columnwidth}{!}{
\begin{tabular}{ll}\toprule
\textbf{Task} & \textbf{Instruction Template} \\\midrule
\multirow{10}{*}{Multilingual Fig-QA} 
    & Given the phrase: '\{\{startphrase\}\}', which of the following is correct?\textbackslash n1. \{\{ending1\}\}\textbackslash n2. \{\{ending2\}\}\textbackslash nAnswer (1 or 2): \\
    & Based on the phrase: '\{\{startphrase\}\}', which answer is more accurate?\textbackslash n1. \{\{ending1\}\}\textbackslash n2. \{\{ending2\}\}\textbackslash nSelect 1 for the first option, or 2 for the second: \\
    & Phrase: '\{\{startphrase\}\}'\textbackslash nWhich of the following best matches the phrase?\textbackslash nOption 1: \{\{ending1\}\}\textbackslash nOption 2: \{\{ending2\}\}\textbackslash nAnswer (1 or 2): \\
    & Here is a phrase: '\{\{startphrase\}\}'\textbackslash nWhich statement is correct?\textbackslash n1. \{\{ending1\}\}\textbackslash n2. \{\{ending2\}\}\textbackslash nChoose 1 for the first, 2 for the second: \\
    & Context: '\{\{startphrase\}\}'\textbackslash nWhich option is true?\textbackslash n1. \{\{ending1\}\}\textbackslash n2. \{\{ending2\}\}\textbackslash nChoose the correct number (1 or 2): \\
    & The phrase is: '\{\{startphrase\}\}'\textbackslash nWhich of the following is correct?\textbackslash n1. \{\{ending1\}\}\textbackslash n2. \{\{ending2\}\}\textbackslash nRespond with 1 or 2: \\
    & For the phrase '\{\{startphrase\}\}', which answer is accurate?\textbackslash n1. \{\{ending1\}\}\textbackslash n2. \{\{ending2\}\}\textbackslash nAnswer 1 if correct, or 2 if correct: \\
    & '\{\{startphrase\}\}'\textbackslash nWhich one of the following is correct?\textbackslash nOption 1: \{\{ending1\}\}\textbackslash nOption 2: \{\{ending2\}\}\textbackslash nChoose 1 or 2: \\
    & Given the statement: '\{\{startphrase\}\}', which of the following is the correct conclusion?\textbackslash n1. \{\{ending1\}\}\textbackslash n2. \{\{ending2\}\}\textbackslash nAnswer (1 or 2): \\
    & Consider the phrase: '\{\{startphrase\}\}'\textbackslash nWhich option is the right one?\textbackslash n1. \{\{ending1\}\}\textbackslash n2. \{\{ending2\}\}\textbackslash nSelect your answer (1 or 2): \\\bottomrule
\end{tabular}

}

\subsubsection{X-CSQA Templates}

\resizebox{\columnwidth}{!}{
\begin{tabular}{ll}\toprule
\textbf{Task} & \textbf{Instruction Template} \\\midrule
\multirow{10}{*}{X-CSQA} 
    & Question: \{\{stem\}\}\textbackslash nChoices:\textbackslash nA. \{\{choice1\}\}\textbackslash nB. \{\{choice2\}\}\textbackslash nC. \{\{choice3\}\}\textbackslash nD. \{\{choice4\}\}\textbackslash nE. \{\{choice5\}\}\textbackslash nAnswer: \\
    & \{\{stem\}\}\textbackslash n\textbackslash nOptions:\textbackslash nA. \{\{choice1\}\}\textbackslash nB. \{\{choice2\}\}\textbackslash nC. \{\{choice3\}\}\textbackslash nD. \{\{choice4\}\}\textbackslash nE. \{\{choice5\}\}\textbackslash nCorrect choice: \\
    & Read the question and select the correct answer:\textbackslash n\textbackslash n\{\{stem\}\}\textbackslash n\textbackslash nChoices:\textbackslash nA. \{\{choice1\}\}\textbackslash nB. \{\{choice2\}\}\textbackslash nC. \{\{choice3\}\}\textbackslash nD. \{\{choice4\}\}\textbackslash nE. \{\{choice5\}\}\textbackslash nAnswer: \\
    & What is the correct answer to this question?\textbackslash n\textbackslash n\{\{stem\}\}\textbackslash n\textbackslash nOptions:\textbackslash nA. \{\{choice1\}\}\textbackslash nB. \{\{choice2\}\}\textbackslash nC. \{\{choice3\}\}\textbackslash nD. \{\{choice4\}\}\textbackslash nE. \{\{choice5\}\}\textbackslash nAnswer: \\
    & Based on the question below, choose the most accurate answer:\textbackslash n\textbackslash n\{\{stem\}\}\textbackslash n\textbackslash nChoices:\textbackslash nA. \{\{choice1\}\}\textbackslash nB. \{\{choice2\}\}\textbackslash nC. \{\{choice3\}\}\textbackslash nD. \{\{choice4\}\}\textbackslash nE. \{\{choice5\}\}\textbackslash nCorrect answer: \\
    & Select the right option for the following question:\textbackslash n\textbackslash n\{\{stem\}\}\textbackslash n\textbackslash nA. \{\{choice1\}\}\textbackslash nB. \{\{choice2\}\}\textbackslash nC. \{\{choice3\}\}\textbackslash nD. \{\{choice4\}\}\textbackslash nE. \{\{choice5\}\}\textbackslash nAnswer: \\
    & Question:\textbackslash n\{\{stem\}\}\textbackslash n\textbackslash nOptions:\textbackslash n1. \{\{choice1\}\}\textbackslash n2. \{\{choice2\}\}\textbackslash n3. \{\{choice3\}\}\textbackslash n4. \{\{choice4\}\}\textbackslash n5. \{\{choice5\}\}\textbackslash nCorrect option: \\
    & Answer the following question by selecting the best choice:\textbackslash n\textbackslash n\{\{stem\}\}\textbackslash n\textbackslash nA. \{\{choice1\}\}\textbackslash nB. \{\{choice2\}\}\textbackslash nC. \{\{choice3\}\}\textbackslash nD. \{\{choice4\}\}\textbackslash nE. \{\{choice5\}\}\textbackslash nYour answer: \\
    & \{\{stem\}\}\textbackslash n\textbackslash nAnswer options:\textbackslash n1. \{\{choice1\}\}\textbackslash n2. \{\{choice2\}\}\textbackslash n3. \{\{choice3\}\}\textbackslash n4. \{\{choice4\}\}\textbackslash n5. \{\{choice5\}\}\textbackslash nChoose the correct one: \\
    & Below is a question and five possible answers. Choose the correct one.\textbackslash n\textbackslash nQuestion: \{\{stem\}\}\textbackslash n\textbackslash nChoices:\textbackslash nA. \{\{choice1\}\}\textbackslash nB. \{\{choice2\}\}\textbackslash nC. \{\{choice3\}\}\textbackslash nD. \{\{choice4\}\}\textbackslash nE. \{\{choice5\}\}\textbackslash nAnswer: \\\bottomrule
\end{tabular}

}

\subsubsection{X-CODAH Templates}

\resizebox{\columnwidth}{!}{
\begin{tabular}{ll}\toprule
\textbf{Task} & \textbf{Instruction Template} \\\midrule
\multirow{10}{*}{X-CODAH} 
    & Which option makes sense?\textbackslash nA. \{\{choice1\}\}\textbackslash nB. \{\{choice2\}\}\textbackslash nC. \{\{choice3\}\}\textbackslash nD. \{\{choice4\}\}\textbackslash nAnswer: \\
    & Select the most reasonable option:\textbackslash nA. \{\{choice1\}\}\textbackslash nB. \{\{choice2\}\}\textbackslash nC. \{\{choice3\}\}\textbackslash nD. \{\{choice4\}\}\textbackslash nAnswer: \\
    & Choose the most logical statement:\textbackslash nA. \{\{choice1\}\}\textbackslash nB. \{\{choice2\}\}\textbackslash nC. \{\{choice3\}\}\textbackslash nD. \{\{choice4\}\}\textbackslash nAnswer: \\
    & Which of the following is the most plausible?\textbackslash nA. \{\{choice1\}\}\textbackslash nB. \{\{choice2\}\}\textbackslash nC. \{\{choice3\}\}\textbackslash nD. \{\{choice4\}\}\textbackslash nAnswer: \\
    & Pick the statement that makes the most sense:\textbackslash nA. \{\{choice1\}\}\textbackslash nB. \{\{choice2\}\}\textbackslash nC. \{\{choice3\}\}\textbackslash nD. \{\{choice4\}\}\textbackslash nAnswer: \\
    & Which statement is most consistent with common sense?\textbackslash nA. \{\{choice1\}\}\textbackslash nB. \{\{choice2\}\}\textbackslash nC. \{\{choice3\}\}\textbackslash nD. \{\{choice4\}\}\textbackslash nAnswer: \\
    & What makes the most sense in this context?\textbackslash nA. \{\{choice1\}\}\textbackslash nB. \{\{choice2\}\}\textbackslash nC. \{\{choice3\}\}\textbackslash nD. \{\{choice4\}\}\textbackslash nAnswer: \\
    & Choose the statement that fits best:\textbackslash nA. \{\{choice1\}\}\textbackslash nB. \{\{choice2\}\}\textbackslash nC. \{\{choice3\}\}\textbackslash nD. \{\{choice4\}\}\textbackslash nAnswer: \\
    & Which option is the most reasonable?\textbackslash nA. \{\{choice1\}\}\textbackslash nB. \{\{choice2\}\}\textbackslash nC. \{\{choice3\}\}\textbackslash nD. \{\{choice4\}\}\textbackslash nAnswer: \\
    & Select the option that best aligns with common sense:\textbackslash nA. \{\{choice1\}\}\textbackslash nB. \{\{choice2\}\}\textbackslash nC. \{\{choice3\}\}\textbackslash nD. \{\{choice4\}\}\textbackslash nAnswer: \\\bottomrule
\end{tabular}

}

\subsection{Topic Classification Templates}
\subsubsection{SIB-200 Templates}

\resizebox{\columnwidth}{!}{
\begin{tabular}{ll}\toprule
\textbf{Task} & \textbf{Instruction Template} \\\midrule
\multirow{10}{*}{SIB-200} 
    & Given the following text, choose the correct category:\textbackslash n\textbackslash n\{\{text\}\}\textbackslash n\textbackslash nCategories:\textbackslash n\{\{options\_\}\}\textbackslash nAnswer: \\
    & What is the topic of the following text?\textbackslash n\textbackslash n\{\{text\}\}\textbackslash n\textbackslash nPossible categories:\textbackslash n\{\{options\_\}\}\textbackslash nAnswer: \\
    & Classify the following text into one of the categories:\textbackslash n\textbackslash n\{\{text\}\}\textbackslash n\textbackslash nCategories:\textbackslash n\{\{options\_\}\}\textbackslash nAnswer: \\
    & Identify the correct category for the text below:\textbackslash n\textbackslash n\{\{text\}\}\textbackslash n\textbackslash nAvailable categories:\textbackslash n\{\{options\_\}\}\textbackslash nAnswer: \\
    & Which category does the following text belong to?\textbackslash n\textbackslash n\{\{text\}\}\textbackslash n\textbackslash nCategories:\textbackslash n\{\{options\_\}\}\textbackslash nAnswer: \\
    & Read the text and select its category:\textbackslash n\textbackslash n\{\{text\}\}\textbackslash n\textbackslash nCategories to choose from:\textbackslash n\{\{options\_\}\}\textbackslash nAnswer: \\
    & Determine the most suitable category for the following text:\textbackslash n\textbackslash n\{\{text\}\}\textbackslash n\textbackslash nCategories:\textbackslash n\{\{options\_\}\}\textbackslash nAnswer: \\
    & What is the correct classification of the following text?\textbackslash n\textbackslash n\{\{text\}\}\textbackslash n\textbackslash nPossible options:\textbackslash n\{\{options\_\}\}\textbackslash nAnswer: \\
    & Choose the category that best describes the following text:\textbackslash n\textbackslash n\{\{text\}\}\textbackslash n\textbackslash nCategories:\textbackslash n\{\{options\_\}\}\textbackslash nAnswer: \\
    & Given the text below, what is its main topic?\textbackslash n\textbackslash n\{\{text\}\}\textbackslash n\textbackslash nPossible categories:\textbackslash n\{\{options\_\}\}\textbackslash nAnswer: \\\bottomrule
\end{tabular}

}

\subsection{Paraphrase Detection Templates}
\subsubsection{PAWS-X Templates}

\resizebox{\columnwidth}{!}{
\begin{tabular}{ll}\toprule
\textbf{Task} & \textbf{Instruction Template} \\\midrule
\multirow{10}{*}{PAWS-X} 
    & Are the following two sentences paraphrases of each other?\textbackslash n\textbackslash nSentence 1: \{\{sentence1\}\}\textbackslash nSentence 2: \{\{sentence2\}\}\textbackslash nAnswer (Yes or No): \\
    & Do these two sentences mean the same thing?\textbackslash n\textbackslash nSentence 1: \{\{sentence1\}\}\textbackslash nSentence 2: \{\{sentence2\}\}\textbackslash nAnswer (Yes or No): \\
    & Determine if the following sentences are paraphrases:\textbackslash n\textbackslash nSentence 1: \{\{sentence1\}\}\textbackslash nSentence 2: \{\{sentence2\}\}\textbackslash nAnswer (Yes or No): \\
    & Paraphrase detection task:\textbackslash n\textbackslash nSentence 1: \{\{sentence1\}\}\textbackslash nSentence 2: \{\{sentence2\}\}\textbackslash nAre they paraphrases? Answer: (Yes or No) \\
    & Do the sentences below convey the same meaning?\textbackslash n\textbackslash nSentence 1: \{\{sentence1\}\}\textbackslash nSentence 2: \{\{sentence2\}\}\textbackslash nYour answer: (Yes or No) \\
    & Decide if the two sentences express the same idea:\textbackslash n\textbackslash nSentence 1: \{\{sentence1\}\}\textbackslash nSentence 2: \{\{sentence2\}\}\textbackslash nAnswer: (Yes or No) \\
    & Check if the two sentences are semantically equivalent:\textbackslash n\textbackslash nSentence 1: \{\{sentence1\}\}\textbackslash nSentence 2: \{\{sentence2\}\}\textbackslash nAnswer: (Yes or No) \\
    & Sentence similarity task:\textbackslash n\textbackslash nSentence 1: \{\{sentence1\}\}\textbackslash nSentence 2: \{\{sentence2\}\}\textbackslash nAre they the same in meaning? Answer: (Yes or No) \\
    & Based on the given sentences, are they paraphrases?\textbackslash n\textbackslash nSentence 1: \{\{sentence1\}\}\textbackslash nSentence 2: \{\{sentence2\}\}\textbackslash nAnswer: (Yes or No) \\
    & Compare the following sentences:\textbackslash n\textbackslash nSentence 1: \{\{sentence1\}\}\textbackslash nSentence 2: \{\{sentence2\}\}\textbackslash nDo they convey the same meaning? Answer: (Yes or No) \\\bottomrule
\end{tabular}

}

\clearpage

\onecolumn
\section{Inference Time Comparison}
\label{sec:performance}
We report the average examples/sec processed for each of the datasets in Table~\ref{tab:speed}. It is important to note that all models are run on a single GPU, except for Meta-Llama-3-70B-Instruct and Llama-2-13B-chat which were run on 4 and 2 GPUs, respectively.

We also present the maximum number of samples each model can handle during inference, alongside the average time taken to process a single batch, all while fully utilizing a single GPU. These results, detailed in Table~\ref{tab:max_bs_time}, provide a clear understanding of each model's efficiency in handling larger batch sizes under optimal GPU utilization.

\begin{table*}[h]
    \centering
    \resizebox{\textwidth}{!}{
    \begin{tabular}{lrrrrrrrr}
\toprule
\textbf{Model} & \textbf{BCOPA} & \textbf{MRPC} & \textbf{FigQA} & \textbf{Amazon Polarity} & \textbf{StoryCloze} & \textbf{YA Topic} & \textbf{Emotion} & \textbf{Avg} \\
\midrule
Qwen1.5-0.5B-Chat & 0.1 & 0.1 & 0.1 & 0.1 & 0.1 & 0.1 & 0.1 & 0.1 \\
phi-2 & 1.4 & 1.4 & 1.4 & 1.4 & 1.5 & 1.4 & 1.4 & 1.4 \\
Meta-Llama-3-70B-Instruct* & 2.9 & 1.3 & 3.2 & 0.9 & 4.9 & 1.1 & 2.1 & 2.3 \\
flan-t5-large & 8.2 & 13.2 & 13.2 & 13.2 & 13.2 & 13.2 & 13.2 & 12.5 \\
Llama-2-13b-chat-hf* & 8.7 & 5.7 & 12.8 & 4.3 & 15.7 & 4.4 & 6.9 & 8.3 \\
Our Approach (roberta-large) & 9.3 & 14.5 & 15.0 & 15.0 & 14.7 & 3.1 & 5.1 & 11.0 \\
bart-large-mnli & 9.7 & 14.1 & 14.0 & 14.2 & 14.1 & 13.7 & 13.8 & 13.4 \\
pythia-6.9b & 12.0 & 0.6 & 4.6 & 0.4 & 0.6 & 2.2 & 0.4 & 3.0 \\
Llama-2-7b-chat-hf & 12.5 & 0.6 & 4.6 & 0.4 & 0.6 & 2.3 & 0.5 & 3.1 \\
Mistral-7B-Instruct-v0.2 & 12.8 & 0.5 & 2.7 & 0.3 & 0.5 & 1.7 & 0.4 & 2.7 \\
pythia-2.8b & 13.6 & 16.7 & 24.9 & 15.2 & 27.2 & 15.1 & 20.9 & 19.1 \\
flan-t5-small & 13.9 & 39.2 & 39.1 & 39.3 & 39.4 & 39.3 & 39.3 & 35.6 \\
Our Approach (roberta-base) & 17.9 & 49.8 & 50.0 & 49.8 & 49.9 & 10.3 & 17.0 & 34.9 \\
\bottomrule
\end{tabular}}
    \caption{The average examples per second processed by each model on each task. * indicates that the model required the use of more than one GPU.}
    \label{tab:speed}
\end{table*}

\begin{table*}[h]
    \small
    \centering
    \begin{tabular}{lcc}
    \toprule
        \textbf{Model} & \textbf{Maximum Batch Size} & \textbf{Mean Inference Time Per Batch (s)} \\
    \midrule
        Qwen2-0.5B-Chat & 240 & 0.0696 \\
        Qwen2-1.5B-Chat & 118 & 0.0580 \\
        aya-23-8B & 36 & 0.3729 \\
        gemma-2-2B & 72 & 0.3473 \\
        gemma-2-9B & 18 & 0.5682 \\
        Meta-Llama-3.1-8B & 36 & 0.2415 \\
        Our Approach (mdeberta-base) & 732 & 0.0270 \\
    \bottomrule
    \end{tabular}
    \caption{The maximum number of samples each model can handle during inference while fully utilizing GPU memory (Nvidia A100 80GB).}
    \label{tab:max_bs_time}
\end{table*}

\end{document}